%% file: main.tex
\documentclass{vldb}
\usepackage{balance}
\usepackage[shortlabels]{enumitem}
\usepackage{multirow}

\usepackage{graphicx}
\usepackage{multirow}
\usepackage{xcolor}
\newtheorem{example}{Example}
\usepackage{graphicx}
\usepackage[labelfont=bf]{caption}
\usepackage{subcaption}
\usepackage{tabularx}
\usepackage{multicol}
\usepackage{booktabs}
\usepackage{color, colortbl}
\setlist[itemize]{leftmargin=*}
\setlist[enumerate]{leftmargin=*}
\definecolor{Gray}{gray}{0.9}
\usepackage{times}

\newcommand{\yuliang}[1]{{\it\small\textcolor{blue}{[[[ {#1}\ --yuliang ]]]}}}
\newcommand{\xiaolan}[1]{{\it\small\textcolor{brown}{[[[ {#1}\ --xiaolan ]]]}}}

\vldbTitle{Deep or Simple Models for Semantic Tagging? It Depends on your Data}
\vldbAuthors{Jinfeng Li, Yuliang Li, Xiaolan Wang, Wang-Chiew Tan}
\vldbDOI{https://doi.org/10.14778/3407790.3407844}
\vldbVolume{13}
\vldbNumber{11}
\vldbYear{2020}

\begin{document}

%\title{Beyond Sentiments: What Else are People Trying to Extract from Online Reviews [Experiments]}
\title{Deep or Simple Models for Semantic Tagging?\\It Depends on your Data}

\numberofauthors{1} %  in this sample file, there are a *total*
% of EIGHT authors. SIX appear on the 'first-page' (for formatting
% reasons) and the remaining two appear in the \additionalauthors section.

\author{
% You can go ahead and credit any number of authors here,
% e.g. one 'row of three' or two rows (consisting of one row of three
% and a second row of one, two or three).
%
% The command \alignauthor (no curly braces needed) should
% precede each author name, affiliation/snail-mail address and
% e-mail address. Additionally, tag each line of
% affiliation/address with \affaddr, and tag the
% e-mail address with \email.
%
% 1st. author
\alignauthor
Jinfeng Li, Yuliang Li, Xiaolan Wang, Wang-Chiew Tan\\
       \affaddr{Megagon Labs}\\
       \email{\{jinfeng,yuliang,xiaolan,wangchiew\}@megagon.ai}
}

\maketitle
\begin{abstract} 
Semantic tagging, which has extensive applications in text mining, predicts whether a given piece of text conveys the meaning of a given semantic tag. The problem of semantic tagging is largely solved with supervised learning and today, deep learning models are widely perceived to be better for semantic tagging. However, there is no comprehensive study supporting the popular belief. Practitioners often have to train different types of models for each semantic tagging task to identify the best model. This process is both expensive and inefficient.

We embark on a systematic study to investigate the following question:
Are deep models the best performing model for all semantic tagging tasks? To answer this question, we compare deep models against ``simple models'' over datasets with varying characteristics. Specifically, we select three prevalent deep models (i.e. CNN, LSTM, and BERT) and two simple models (i.e. LR and SVM), and compare their performance on the semantic tagging task over 21 datasets. 
%The results showed that deep models are not %necessarily better than simple models when the %characteristics of datasets are variable. %\xiaolan{Maybe say they did not perform well on large datasets first. The previous sentence claims they are not better, so the next sentence better elaborate over this point.} \xiaolan{Maybe add a transition sentence? e.g., To better understand their performance} 
%To understand what are the exact characteristics of datasets affecting tagging quality, we selected the representative simple model (SVM) and deep model (BERT), and compared their performances on different types of datasets. 
Results show that the size, the label ratio, and the label cleanliness of a dataset significantly impact the quality of semantic tagging. Simple models achieve similar tagging quality to deep models on large datasets, but the runtime of simple models is much shorter. Moreover, simple models can achieve better tagging quality than deep models when targeting datasets show worse label cleanliness and/or more severe imbalance. Based on these findings, our study can systematically guide practitioners in selecting the right learning model for their semantic tagging task. %\textcolor{blue}{In particular, we found that simple models outperform deep models on larger datasets with higher label ratios or worse label cleanliness, when runtime is a concern}. 
%The conditional outperformance of deep models %suggests that practitioners should carefully %select learning models when they aim to achieve %the best tagging quality. 
%Our results will systematically guide practitioners in selecting the right learning model for their semantic tagging task.
%
%To further assist practitioners to pick right %learning models, we generated a comprehensive heat map that compares tagging qualities across varied combinations of models and datasets. The heat map will be an instructional reference for practitioners to adopt appropriate models and set accurate expectations when tagging sentences for their own datasets.

%\wctan{heat map or table?} \jinfeng{I prefer heat map that uses colors and reflects the effect of data characteristics on F1}

%\textcolor{red}{Our results indicate that practitioners should pay attention to dataset characteristics when they apply deep models for better \tagging{} quality. We also generate a comprehensive heat map from our results that can help practitioners to adopt appropriate models and set expectation on tagging performance for their datasets}.  

\end{abstract}

\input{introduction.tex}

\input{apps.tex}
\input{algorithm.tex}

\input{data.tex}

\input{experiment.tex}

%\input{discussion.tex}
\input{conclusion.tex}
\input{acknowledgement.tex}

%\balance
\bibliographystyle{abbrv}
\bibliography{main}

\input{appendix.tex}

\end{document}

%% file: introduction.tex
\section{Introduction} \label{sec:intro}
A lot of applications for processing text rely on tagging words, phrases or sentences with semantically informative tags.
%} \nikita{more straightforward?}\jinfeng{Origin: ``A lot of applications for processing text rely on semantic tags that are annotated on the text.''}
Sentiment analysis~\cite{2012Liusynthesis, bingliusaom, pontiki2016semeval}, for example, annotates sentences or phrases with a sentiment tag that indicates whether the sentence has a positive or negative sentiment. These sentiment tags are exploited by downstream applications to determine appropriate actions. Another example is entity tagging, which determines if a span in the text refers to a real-world object. 
%\textcolor{purple}{
Generally speaking, the task of annotating text with semantic tags can be referred to as the semantic tagging problem.
%} \jinfeng{Origin: ``We refer to the problem of annotating text with semantic tags as the {\em semantic tagging} problem.''}
More precisely, semantic tagger takes a piece of text and a predefined tag as inputs, and outputs whether this text conveys the semantics of the tag. %\textcolor{purple}{
In this paper, we focus on short text, which can be a sentence, a paragraph, or a passage. We also refer short text loosely as sentence.
There are two types of methods for semantic tagging: rule programming and supervised learning. Rule programming-based methods require an expert to specify rules for semantic tagging. This is often error-prone and requires significant programming effort. 
% since it is difficult to specify a good set of rules for semantic tagging in general. 
In contrast, supervised learning models do not require much programming effort. However, training these models requires labeled data but can typically produce models with good semantic tagging results.
%\wctan{do we have evidence of less used or frequently used? I removed the mentions of these claims.}
%\wctan{rewrote. pls read} \jinfeng{It reads great!}

Our focus in this paper is on supervised learning models.
Deep learning models (or deep models in short) have become popular methods for semantic tagging today. One reason why deep models are popular for semantic tagging is that they are often more capable of learning complicated functions than other kinds of models. Another reason is that the superiority of deep models has been reported by many publications. For example, deep models achieve good prediction quality that is close to the human prediction on GLUE SST-2 sentiment classification task~\cite{wang2018glue, gluebenchmark}. Some recent studies~\cite{NegiB15emnlp, NegiAMB16starsem, WanMNM19acl, GuyMNR17www, WangPWZ19msr, NovgorodovEGR19www, MoralesZ17emnlp, HuaNBW19naacl, StabMSRG18emnlp, WeberUG12wsdm} made comparisons between deep models and simple models (i.e., machine models that do not leverage deep learning) to understand whether deep models are always superior to simple models. 
%\wctan{what tasks are compared in these studies?}\jinfeng{other semantic tagging tasks. Added one sentence below.} 
They conducted comparisons on various tagging tasks such as suggestion mining~\cite{semeval19NegiDB} or humor detection~\cite{PotashRR17semeval}. Their results reveal marginal or sometimes no improvements of deep models over simple models.  
It is therefore natural to ask whether
deep models are better than simple models when developing solutions for semantic tagging.

%\textcolor{purple}{
Semantic tagging forms the core of many tasks including sentiment classification, suggestion mining, and humor detection. Existing studies, however, compare deep and simple models only on individual tasks. Furthermore, they do not provide insights on how dataset characteristics affect the performances of different models. Consequently, it is hard to generalize their model selection criteria to new tasks or new datasets for the same task. Hence, given a new dataset, it is still unclear whether selecting a deep model will bring the best tagging performance.
%}
% The goal is to identify the model which achieves best performance for the targeted task. These studies do not consider dataset characteristics when selecting the best model. As a consequence, no previous studies show how to generalize the model selection to new tasks and even new datasets of the same task. They developed selection guidance at the granularity of individual task but not at the granularity of individual dataset characteristic. Given a new dataset, it is still unclear whether selecting deep models can bring the best semantic tagging performance.}
%\jinfeng{add one paragraph to motivate 21 datasets and our study}

% select five models and 21 different datasets
In this paper, we embark on a systematic study to understand the performance tradeoffs of deep models vs. simple models for semantic tagging. Towards this goal, we selected 3 representative deep models: CNN, LSTM, and BERT and 2 representative simple models: LR and SVM. CNN~\cite{Kim14emnlp} and LSTM~\cite{HochreiterS97neco} are well-known methods that have been widely used in both the academic and industry communities and more recently, BERT~\cite{devlin2018bert}. 
%represents the most up-to-date method that w%on the best paper award in NAACL 2019. 
To make a meaningful comparison and systematic study, we collected 21 real datasets that are frequently used in semantic tagging. These datasets exhibit several prominent data characteristics, including (1) a variable number of labels (thousands to millions); (2) a wide range of tag-conveying label ratio (1.6\% to 71.4\%, ratio $< 25\%$ considered as imbalance in this paper); (3) different label cleanliness (clean and dirty labels).

We evaluate the quality of semantic tagging on five selected models on 21 datasets and we obtain a rather surprising finding.
%that overturns the general perception %towards deep models. 
We find that deep models and simple models are complementary to each other on the task of semantic tagging. Specifically, deep models perform significantly better on smaller datasets, while simple models can be trained more efficiently on larger datasets and achieve similar semantic tagging quality. Therefore, one should select deep models or simple models for semantic tagging based on the actual dataset characteristics and requirements on efficiency. %On large datasets that contain more than 100,000 labels, deep models only achieve a maximal 0.03 F1 superiority yet takes days for training. If training time is a concern, deep models are not better choices than simple models for large datasets.

%Deep models can offer better tagging quality, but they alone are inadequate to guarantee satisfactory tagging quality. In our experiments, the tagging F1 of the best deep model varies from 0.96 to 0.15. This is because the tagging performance is at the same time regulated by the dataset characteristics, including the number of labels, the ratio of tag-conveying labels, and the label cleanliness. We conducted further experiments and confirmed that more labels, a higher ratio, and better label cleanliness contribute to greater tagging quality of deep models. 

Based on our findings, we develop a comprehensive heat map to guide practitioners on selecting the appropriate model for the desired semantic tagging performance for their datasets. This heat map shows the characteristics of the datasets and their quality score of semantic tagging with different tagging models. By using this heat map, practitioners can estimate the semantic tagging quality gain while adopting different deep or simple models. At the same time, they can also try to improve the dataset characteristics to improve the quality of semantic tagging. \\

%To help practitioners select appropriate models and set expectation on the tagging performance for their datasets, we prepare a comprehensive heap map that shows the characteristics of the 21 datasets and their tagging F1's with BERT and SVM. By using this heap map, practitioners can estimate the tagging quality gain of adopting deep models. Meanwhile, they can try to improve the dataset characteristics to push the tagging quality to higher upper limits. \\

\noindent\textbf{Contributions.} We systematically evaluate deep models and simple models for the task of semantic tagging.
%We selected X models, curated Y datasets of varying characteristics, and conducted a series of experiments investigate the performance of the models in these two models and datasets. 
%\wctan{Are we releasing our collection of datasets and models?} \jinfeng{Yes. I added one sentence about it to the end of this paragraph. Will work on the release later.}
Our key contributions are as follows. (1) We surveyed a number of applications to motivate our study. We selected three representative deep models and two simple models that are widely used to develop these applications. We collected 21 datasets of varying characteristics for a comprehensive study. (2) We conducted extensive evaluations to obtain performance of semantic tagging of the five selected models on all the datasets. We found deep models do not necessarily perform better than simple models on large datasets. (3) We evaluated the effects of dataset characteristics on the quality of semantic tagging. We found the training size, label ratio, and label cleanliness impact the quality of semantic tagging. (4) We generated a comprehensive heat map that can guide practitioners to decide whether they should adopt deep models or simple models and anticipate the performance of semantic tagging for their datasets. To facilitate future research, we will release our collection of datasets, models, and implementations at https://github.com/rit-git/tagging. \\

\noindent\textbf{Outline.} We survey a number applications in Section~\ref{sec:apps}. We discuss the designs of selected deep models and simple models in Section~\ref{sec:models}. We introduce the collected datasets and their characteristics in Section~\ref{sec:dataset}. We perform experimental evaluations and comparisons in Section~\ref{sec:experiment}. We analyze the effect of dataset characteristics and present key findings in Section~\ref{sec:effect_characteristic}. We conclude our study in Section~\ref{sec:conclusion}.

%% file: apps.tex
\section{Latest applications of semantic tagging} \label{sec:apps}
%\section*{APPENDIX: LATEST APPLICATION of SEMANTIC TAGGING} 

%\yuliang{To Jinfeng: provide examples of these types of sentences in the hotel domain.}\jinfeng{We need to delete something to make a space} \xiaolan{Change the section name?} \jinfeng{Changed}

% suggestion, 
% content, (spoiler)
% hilarious snippet 
% argument
% answer

%When a merchant holds plentiful online reviews, a natural question is how to generate value from these data. Most of the existing research focuses on sentiment analysis to un- derstand user’s opinion on the business. In recent years, more and more applications emerge in the field making use of review data from broader perspectives. In this section, we will present what sentences are deemed attractive and summarize these new applications.

% People extensively extract sentences from online reviews. One classic example is sentiment analysis, where substantial efforts focus on extracting positive or negative comments to develop opinion mining applications. In recent years, a number of new applications emerge in the field attempting to extract sentences from a broad perspective. In this section, we will present six types of emerging sentences extraction and their applications. 
%We summarized related papers at Table~\ref{table:papersummary} \\

Semantic tagging has broad applications in the field. In this section, we focus on introducing the latest ones. These novel applications come from publications and industrial practices in recent three years. We investigated related literature and link them to five new tags, i.e. Tip, Product Description, Humor, Spoiler, and Argument. These new tags and applications can contribute more ideas to practitioners to leverage semantic tagging on their datasets. % new

\subsection{Tip}
 
%\yuliang{combined tips and suggestions. Still need to fix the flow and shorten the length.} \jinfeng{I have combined further based on your version}

% suggestion definition
%Reviewers often provided suggestions in their comments. They gave constructive plans that can enhance customer satisfaction based on their actual experiences. Useful suggestions are deemed explicitly actionable~\cite{NegiB15emnlp}, i.e. they expressed actions that are easy to follow. For example, a hotel review may recommend visiting a nearby attraction for beautiful scenery or going to a local canteen for delicious food. Some suggestions help avoid negative feelings. One such example may be a warning to females that it is not safe to walk alone on the street late at night. 

Experience-sharing websites such as TripAdvisor and Yelp encourage users to write short yet useful tips such as those in Table~\ref{table:tripadvisor_tip} and Table~\ref{table:yelp_tip}. This motivates the tip tagging task, which automatically identifies tip-expressing sentences from reviews instead of mandatorily asking users to do so. In comparison to a full review, which is sometimes lengthy, a tip is much shorter yet still reveals the most prominent features of targets.

A tip usually contains suggestive information that enables future customers to improve their consumption experience. For example, a tip may recommend paying 20\% tips for services in Washington DC or ask for English menus in Hong Kong as in Table~\ref{table:tripadvisor_tip}. Guy et al.~\cite{GuyMNR17www} studied tagging practical and short texts from reviews. The authors noticed that readers might feel overwhelmed by reading massive reviews and tend to miss key points. Also, a lengthy review is not friendly to cell phones as the screens have limited space. Wang et al.~\cite{WangPWZ19msr} explored how to highlight tip sentences in software review to help developers pinpoint potential improvements of the software. The developers feel easier to identify emerging issues or prominent problems by reading tips. Weber et al.~\cite{WeberUG12wsdm} analyzed queries sent to Yahoo Search and found a large fraction of queries containing how-to intent, such as ``how to zest a lime without zester''. They tagged tip answers from Yahoo!Answer databases and showed the tip answer at the front of the screen when receiving a how-to query from the user. The conciseness and practicability of a tip help users obtain answers quickly. Zhu et al.~\cite{ZhuLZ18dss} observed that a practical tip tends to mention the speciality of a service, i.e. aspects frequently discussed in reviews of this service but less frequently in reviews of other services. 

%Sapna et. al.~\cite{NegiB15emnlp} asked human annotators to label explicit customer-to-customer suggestions and provided two guidelines. First, a suggestion sentence must explicitly state the intention of giving a recommendation. Second, the suggestion should have the potential benefiting customers. For example, a suggestion sentence may look like 'Try the cup cakes at the bakery next door'. An opposite example would look like "the cup cakes from the bakery next door were delicious" as it does not indicate the suggestion explicitly. 

A number of studies focus on tips that give suggestions explicitly. Sapna et al.~\cite{NegiB15emnlp} asked human annotators to label explicit customer-to-customer suggestions that meet two conditions: (1) explicitly state the intention of giving a recommendation; (2) have the potential to benefit customers. For example, a suggestion sentence to be labeled may look like 'Try the cup cakes at the bakery next door'. An opposite example may look like "the cup cakes from the bakery next door are delicious". Since suggestion-expressing sentences are valuable, tagging them from massive reviews becomes an essential task for SEMEVAL 2019 competitions~\cite{semeval19NegiDB}. To perform this task, competitors built classifiers to predict whether a sentence expresses a suggestion or not. The task provided 9092 and 808 labeled sentences in software and hotel domain correspondingly. The best performing algorithm~\cite{semeval19liuWY} adopted the recent pre-trained BERT~\cite{devlin2018bert} model that can work decently with a small number of training data.

\setlength{\tabcolsep}{5pt}
\begin{table}[t]
\caption{TripAdvisor city guide tips}
\label{table:tripadvisor_tip}
\centering
\scalebox{0.8}{
\begin{tabular}{lp{0.8\linewidth}}
\toprule
City  & Tips  \\ 
\midrule
\multirow{2}{*}{Washington DC} & 1. 20\% tip is customary for most services \\[+1pt]
  & 2. Avoid walking through parks at night \\[+1pt]
\midrule

\multirow{2}{*}{Hong Kong} & 1. Many 'local' restaurants offer menus in English. Ask if they have one before you sit down. \\ [+1pt]
  & 2. Grab an Octopus Card to store money and use on buses, trains and in convenience stores (you have to pay deposit). It will save you a lot of time queuing or fishing for change. \\
\bottomrule
            
\end{tabular}
}
\end{table}

\setlength{\tabcolsep}{5pt}
\begin{table}[t]
\caption{Yelp tips}
\label{table:yelp_tip}
\centering
\scalebox{0.85}{
\begin{tabular}{lp{0.8\linewidth}}
\toprule
Business  & Tips  \\ 
\midrule
\multirow{2}{*}{Restaurant} & 1. There aren't many sushi choices in Mountain View, so take what you can get. \\[+1pt]
 & 2. Free Parking in the lot across the street and around the residential areas. \\[+1pt]
\midrule

\multirow{2}{*}{Coffee shop} & 1. only place in Bay Area I've been able to find soft boiled eggs \\ [+1pt]
  & 2. Delightful cafe with solid pastries. Drinks left more to be desired and service is a bit slow, so make sure you have ample time!. \\
\bottomrule
            
\end{tabular}
}
\vspace{-3mm}
\end{table}

\vspace{3mm}
\subsection{Product Description}

E-commerce websites such as Amazon and eBay collect a lot of customer reviews in recent years. Among these customer reviews, product description is one of the most informative sentences. Product description from customer reviews often contains supplementary information that is missing in official product descriptions. These customer product descriptions are attractive to people due to multiple reasons. First, they come from independent users who have no personal connections with merchants. Second, they summarize crucial facets concerning actual user experience. Third, they introduce exceptions to avoid unnecessary orders when the product is not applicable. For example, a buyer posted actual experience after using a camera for several weeks. He described the actual performance of the camera, such as the photo quality, the ergonomics design, and the battery duration. He might also describe the returning policy and experience if he did not like the camera. Therefore, other customers can determine whether to buy a product according to these customer reviews.    

% why useful
Tagging product description from reviews serves critical purposes. A study~\cite{NovgorodovEGR19www} about eBay reported that official descriptions were always lacking for new items (e.g., fashion products). For old items, details were also frequently missing from official descriptions. To solve this problem, the authors tagged user descriptions from customer reviews and conducted a user study to see whether customer descriptions could provide more information. The user study showed that people believed customer descriptions were more informative and objective than sentimental sentences such as ``It was the best socks ever''. One example of how authors tagging customer descriptions is shown in Example~\ref{eg:prodep}. Some other studies also showed how to tag customer descriptions and apply them to industrial products. Rakesh et al. ~\cite{RakeshDARSR18www} proposed to tag and organize customer descriptions according to different aspects of a product. They further presented a review visualizing tool empowered by product-describing sentences based on their proposal. Mitcheltree et al. ~\cite{airbnb18naacl} distilled aspect-describing sentences of a user and used these sentences to understand the tastes of this user. Zhang et al. ~\cite{ZhangQAFFR19www} detected defects of a product from customer reviews and provided instructions for customers when using the corresponding product. These studies have substantially simplified the purchase decisions for customers. 

User-written product descriptions help merchants and readers better understand sentimental descriptions. This is especially important as pure sentiment words only provide none objective information. Furthermore, merchants expect users to describe products objectively instead of expressing sentiments that might expel potential buyers. On the other hand, customer readers also look for objective evidence to avoid being misled by sentiments. Therefore, customer descriptions are complementary to sentimental comments and necessary for customers to choose suitable products. 

\begin{example}
\label{eg:prodep}
Perfect thickness for shoes or boots. Extra padd-\\ing at toes. The quality is excellent. Easy to handle and very comfortable. No shrinkage in washer or dryer.
\end{example}

\subsection{Humor}
Humor has the power to bring people happiness. YELP~\cite{yelp_challenge} started exploring humor in customer reviews and offered user interfaces to label whether a review is funny or not. Example~\ref{eg:humor:clean} and~\ref{eg:humor:good} show two examples of commenting ``clean room'' and ``delicious buffet''.  Although informative online reviews can help us make decisions, reading, and digesting these reviews are stressful and unpleasant. Different from others, a humorous review can bring happiness to the readers, even it can not provide more information. Therefore, humorous reviews can attract more attention from users for those apps who prioritize them than other reviews. 

Tagging humorous text from a corpus attracts a broader discussion recently. Yang et al. ~\cite{YangLDH15emnlp} studied humor extraction that classifies whether a given text contains humor or not. They identified four types of features that could better classify humor. Morales and Zhai~\cite{MoralesZ17emnlp} tried to tag humor from merchant reviews as the YELP Challenge released a large annotated corpus. Each of the 6,685,900 reviews contains the number of votes from readers who thought the review was funny. Cattle and Ma ~\cite{CattleM18coling} emphasized that a humor sentence expresses semantics that goes beyond human expectation. Thus humor-expressing sentences tend to contain words that are semantically different. Blinov et al. ~\cite{BlinovBB19acl} prepared a dataset of funny dialogues in Russian. They calculated the length distribution of both jokes and non-jokes and found that a text with more than 100 characters is likely to be a joke than non-joke. Yu et al. ~\cite{YuTW18acl} studied how to generate pun with neural networks. Due to the prevalence of humor, SEMEVAL 2017 held two tasks on humor detection~\cite{MillerHG17, PotashRR17semeval}. These studies reveal the essence of humor and also the unique features of humorous comments among all types of sentences.

\begin{example}
\label{eg:humor:clean}
Restaurant so clean....you can eat off the floor!!!!
\end{example}

\begin{example}
\label{eg:humor:good}
I usually am NOT a fan of buffets.. but this place really puts other buffets to SHAME!
\end{example}

\subsection{Argument} %reason
% explanative sentences
% whether it is an argument
% supporting or opposing argument
% type of argument

% what is arguments, and why important, types
Arguments provide evidence to support or oppose an opinion and are useful for explanations. Compared with pure sentimental comments such as ``the room is uncomfortable'', argument sentences usually contain factual details such as ``the room has a smell of mold''. Since arguments are generally different opinions towards the same proposition, they often bring much information from wide perspectives. Example~\ref{eg:argument:nuclear} and~\ref{eg:argument-non:nuclear} show how argument and non-argument sentences pointing to the proposition ``nuclear energy is good'' from questia~\cite{questia}. These two examples demonstrate that argument sentences provide more explanatory information towards a proposition.

Argument mining studied automatic tagging of argument expressing sentences that had caused extensive discussion in review mining. Hua et al.~\cite{HuaNBW19naacl} started exploring argument tagging in the domain of paper reviews, as they noticed paper reviewing takes approximately 63.4 million hours in 2015. They obtained public reviews from ICLR and extracted argument sentences from these reviews. Then they classified the argument sentences into five classes (i.e. request, evaluation, reference, quote, or fact) to better understand peer comments. Stab et al. ~\cite{StabMSRG18emnlp} studied argument mining for general text. They collected datasets from Questia and annotated arguments for eight controversial-topics such as abortion and marijuana. Kim and Hovy~\cite{KimH06acl} started mining reasons depicting the pros and cons of the targeted service or product. Poddar et al. ~\cite{PoddarHL17emnlp} applied argument mining to customer reviews to tag supporting opinions for a given aspect. These studies provide us more knowledge of argument sentences and show us their potential applications in actual products.

% existing practices on argument
\begin{example}
\label{eg:argument:nuclear}
The entire nuclear fuel chain generates lots of long-lasting radioactive waste. 
\end{example}

\begin{example}
\label{eg:argument-non:nuclear}
Can truly green, renewable sources of energy replace nuclear power?
\end{example}

\subsection{Spoiler}
% alert for readers or for writers
% what you shouldn't write generalize the idea a bit. 
Spoiler commonly exists in the reviews of media works such as TV series. In such reviews, audiences express their evaluations and ideas towards media works, while they inevitably quote specific plots to support their ideas. These spoilers might ruin the expectation and enjoyment of audiences who have not watched the media works. Therefore, a spoiler alert is necessary to warn readers before they browse the reviews. For example, the movie encyclopedia website IMDB mandatorily requires reviewers to add an alert message to the title if they are going to include spoilers in the comments. Online wiki Lostpedia also banned spoilers since mid-2008. Due to the broad necessity of spoiler alerts, 
automatic detection becomes more desirable so that reviewers can avoid writing spoilers, and readers can avoid reading spoilers.

Spoiler detection is a field with little exploration, mainly due to the shortage of datasets. There are only two open-source datasets for spoiler detection. One is TV Tropes dataset released by Boyd-Graber~\cite{Boyd-GraberGZ13asist}. This dataset contains spoiler sentences that disclose important TV show plots, such as the ending of main characters. An example of TV Tropes is shown in Example~\ref{eg:spoiler:tv}. The other one is a large dataset consisting of millions of book spoiler sentences released by Wan et al. ~\cite{WanMNM19acl}. These authors found that spoiler sentences tend to appear in the later part of a review and book spoilers are much variable in words correspondingly to the book-specific contents, such as characters' names. Due to the ubiquitousness of spoilers and the necessities of automatic detectors, this paper received a number of media reports. An example of book spoilers is shown at Example~\ref{eg:spoiler:book}. 

%tends to contain the characters' names and spoilers of different books vary a lot in the words used to describe book-specific contents.

\begin{example}
\label{eg:spoiler:tv}
None of the Harmons survive.
\end{example}

\begin{example}
\label{eg:spoiler:book}
This book had all the potential to be great: a political thriller with tons of twists  including, but not limited to, killing off the main character in the middle of the book .
\end{example}

%% file: algorithm.tex
\section{Representative models} \label{sec:models}

In the previous section, we saw a sentence can have five types of tags. For each type, a binary classifier can be trained using example data. We first formally define the semantic tagging problem and describe the common training pipeline. We then describe the representative sentence classification models for semantic tagging.

\subsection{The Semantic Tagging Problem}
%formulate sentences ion as binary classification
The semantic tagging problem is generally solved by a binary text classification solution. The solution enables people to classify a text into one of the two classes: $\{\text{\it desired, not-desired}\}$. Based on the classification results, the tagged sentences are essentially those classified as ``{\it desired}''.

The sentence classification problem is widely studied as text classification~\cite{KowsariMHMBB19information, AggarwalZ12bmining2012} %\xiaolan{add citations} \jinfeng{Added a survey and text classification} 
in the machine learning and natural language processing community.  %There are two generic types of solutions for this problem: \textit{unsupervised} solution, which does not require any labeled data, and \textit{supervised} solution, which often requires an extensive number of labeled data. Although unsupervised solution does not require any labeled data, it often requires domain-experts to write domain-specific rules based on extensive experience. Therefore, in this paper, we mainly focus on the supervised solution.
To solve the supervised sentence classification problem, people generally perform experiments at three steps: 
(1). label data preparation; (2). input sentences representation; and 
(3). model selection. The quality of each step can significantly affect the accuracy of the classification.

\vspace{2mm}

\noindent \textbf{Label data preparation. }
%label preparation
The first step is label preparation. A label instance is in the format of $(text, label)$, where $text$ is the raw sentence, and $label$ is either 1 (positive training example) or 0 (negative example). After we have collected a number of labels, we apportion the data into training and test sets, with a pre-specified split (e.g., 80-20). 
The purpose of splitting data is to mimic actual prediction, where the model training is performed on the training set but the model evaluation is conducted on the testing set.  

\vspace{2mm}

\noindent \textbf{Input sentence representation. } The second step is pre-processing that raw sentences are converted into numeric representations by classification models. At this step, a sentence is tokenized into words. The purpose of tokenization is to represent each word using a numeric word vector. By aggregating word vectors, we can obtain the final numeric representation of the sentence. There are two commonly used sentence representation methods: Bag-of-words (BoW) and Word embeddings. The choice of a method heavily depends on the choice of the classification model.

\vspace{2mm}

\noindent \textbf{Model selection. } Classification models include two sub-categories: statistical models and neural network models. Compared with neural network models, statistical models are usually simpler (smaller in size) and less computationally intensive. However, statistical models require more domain knowledge in constructing a set of essential features for each specific task. In contrast, 
neural network models are larger in size (more parameters) and more computationally intensive (usually trained on hardware accelerators like GPUs), but they do not require feature engineering and can learn how to represent the text directly from training data. Due to this trade-off, people often select different models for different scenarios.

However, we are witnessing a rise in the popularity of neural network models, such that they are considered best performing models for all semantic tagging tasks. We revisit this popular belief and compare an array of statistical models (Logistic Regression (LR) and Support Vector Machine (SVM)) and deep models (Convolutional Neural Network (CNN)~\cite{Kim14emnlp}, 
Long Short-Term Memory (LSTM)~\cite{HochreiterS97neco}, Bidirectional Encoder Representations from Transformers (BERT)~\cite{devlin2018bert}).

% Along with recent advances in deep learning, 
% it is widely believed that neural network models can significantly improve 
% the tagging accuracy, in comparison to statistical models.
% To validate this ``belief'', %we conduct a comprehensive comparison over these two categories of models. 
% we select Logistic Regression (LR) and Support Vector Machine (SVM) as the representative statistical models (or non-deep models), and Convolutional Neural Network (CNN)~\cite{Kim14emnlp}, 
% Long Short-Term Memory (LSTM)~\cite{HochreiterS97neco}, and Bidirectional Encoder Representations from Transformers (BERT)~\cite{devlin2018bert} (Section~\ref{sec:deep}) as the representative 
% neural network models (or deep models) for further analyses.
\subsection{Simple Models (LR/SVM)} \label{sec:nondeep}
We use Bag-of-words (BoW) as the input representations for simple models. First, BoW splits a text into tokens, each of which can be a unigram (a single word) or a bigram (two consecutive words). In our experiments, we found a combination of unigram and bigram yields the best tagging quality. Second, BoW calculates the number of distinct tokens as vocabulary size (denoted as $d$), so that each token has a unique position in a $d$-dimensional vector. Third, BoW represents a text as a $d$-dimensional vector, where the $i$-th position records the count of the corresponding token in the text. So there will be many 0 in the vector if the vocabulary size $d$ is large.

BoW also uses IDF (inverse document frequency) to weigh each of the $d$ tokens. IDF assumes that a token is more important if it appears in less number of sentences, and the formula of IDF is $idf(t) = log [ n / df(t) ] + 1$, where $t$ is the corresponding token, $n$ is the total number of sentences, and $df(t)$ is the number of sentences that contain $t$. For example, ``My advanced geometry class is full of squares'' is a pun-containing humorous sentence, in which the word ``my'' is less informative than the word ``geometry'' and ``squares''. \\

%class weight

\noindent\textbf{Logistic Regression (LR).} A classical simple model is LR~\cite{friedman2001elements}, which is derived from linear regression and uses Sigmoid~\cite{friedman2001elements} function to scale the output into a real number between 0 and 1. This real number indicates the probability of an instance belonging to a class. LR is widely used in production to analyse large-scale data~\cite{harrell2015regression, BerryL97lib}. It has the advantages of low computational overhead and high parallelism~\cite{HugginsCB16nips, ZahariaCDDMMFSS12nsdi}. \\

\noindent\textbf{Support Vector Machine (SVM).} Another popular simple model is SVM~\cite{suykens1999least}, which separates two classes by their border points. We therefore use linear SVM, which adopts a straight line, to separate different classes. Linear SVM is known to be empirically effective in high-dimensional space, where data become sparse and tend to be linearly separable~\cite{suykens1999least, svm98ecml, PradhanWHMJ04naacl, svmchemistry}. Besides, linear SVM scales to large number of labels better than its variants of using non-linear kernels in training time~\cite{sklearnlinearsvm, ChangHCRL10jmlr}.

\subsection{Deep Models (CNN/LSTM/BERT)}\label{sec:deep}
\vspace{2mm}
\noindent \textbf{Convolutional Neural Network (CNN).} CNN~\cite{Kim14emnlp} has traditionally been used for image classification but is now used for several NLP tasks. CNN tokenizes a text into unigram words. Each word is represented with a pre-trained $k$-dimensional vector. Two words will have similar word vectors if they are semantically similar (e.g., synonyms). The text is represented as a $m \times k$ matrix, where $m$ is a predefined maximum sequence length. In this paper, we also name the $m \times k$ matrix as feature matrix. CNN consists of multiple convolutional layers and pooling layers. 
The convolutional layers convert the consecutive elements (e.g., bi-grams like ``please add'')
of the input matrix into a sequence of feature vectors with a sliding window.
Then the pooling layers aggregate the sequence into a shortened one
via a max aggregate function. The convolutional and pooling layers
are usually stacked to obtain a single vector that represents the sentence meaning.
Intuitively, when CNN is applied on text sequences, the neural network first constructs
low-level features from local information like bi-grams and tri-grams, and then
constructs higher-level features (e.g., whether a span contains related information) 
from the low-level ones.

\vspace{2mm}
\noindent \textbf{Long Short-Term Memory (LSTM).} LSTM~\cite{HochreiterS97neco} is based on recurrent neural network (RNN)~\cite{Goodfellow-et-al-2016}. It has been shown to be effective for several tasks including classification, tagging, and translation. LSTM uses the same input representation as CNN, i.e. a matrix of $m$ word vectors. However, unlike CNN, LSTM sequentially (left to right) processes the text over time and keeps its \emph{hidden state} through time. The hidden state can capture any meaningful features that appeared in the prefix of the text up to the current timestamp. This enables LSTM to capture arbitrary long-term dependencies. Compared with vanilla RNN, LSTM (and its variants like GRU \cite{chung2014empirical}) partially solves the gradient exploding
and vanishing problem that results in improved model performances with 
its specially designed components (i.e. input/output/forget gates).
%\yuliang{added CNN and LSTM. Pls check.}\jinfeng{Checked. I changed a few words}

\vspace{2mm}
\noindent \textbf{Bidirectional Encoder Representations from Transformers \\  (BERT).}
% BERT~\cite{devlin2018bert} emerges as one of the most effective models and won the best paper award of NAACL-HLT 2019. 
BERT~\cite{devlin2018bert} is an award-winning state-of-the-art model. Similar to CNN and LSTM, BERT also uses a matrix of word vectors to represent a text. BERT applies attention technique to represent a text with weighted word vectors, such that relevant tokens have higher weights than irrelevant ones (e.g. ``great'' and ``delicious'' as relevant tokens and ``that'' as irrelevant token in YELP polarity classification). BERT derives its performance from language representation pre-trained on a large corpus, Wikipedia. So BERT can learn a default vector and the weight of a word in the Wikipedia corpus. When applying BERT to domain-specific data, these word vectors and weights are optimized according to labels, such that domain-specific knowledge can be incorporated.

%% file: data.tex
\setlength{\tabcolsep}{6pt}
\begin{table}[!t]
\caption{Statistics of 21 Dataset. We obtained 19 datasets in 5 different application domains. We further created two additional large datasets from FUNNY and BOOK by balancing their positive and negative labels. Among all 21 datasets, 6 of them are large datasets with more than 100K records (highlighted in gray). The labels of FUNNY, BOOK, FUNNY*, and BOOK* are dirty due to the existence of missing annotations.}
\label{table:datasets}
\centering
\small
\scalebox{0.9}{
\begin{tabular}{lcccc} \toprule
\textbf{Dataset} &\textbf{Application} & \textbf{\#Record}   & \textbf{\%Positive} & \textbf{Vocabulary} \\ \midrule    
\multicolumn{5}{c}{\textbf{Original Datasets}} \\\midrule
SUGG~\cite{semeval19NegiDB}& Tip    & 9,092       & 26.2\%      & 10K \\ %\hline
HOTEL~\cite{NegiB15emnlp}  & Tip  & 7,534        & 5.4\%       & 7K \\ %\hline
SENT~\cite{WangPWZ19msr}   & Tip   & 11,379        & 9.8\%      & 8K  \\ %\hline
PARA~\cite{WangPWZ19msr}   & Tip     & 6,566        & 16.8\%     & 8K   \\ %\hline
\rowcolor{Gray}
FUNNY~\cite{MoralesZ17emnlp}   & Humor  & 4.75M  & 2.5\%        & 571K  \\ %\hline
HOMO~\cite{MillerHG17}   & Humor    & 2,250        & 71.4\%     & 5K   \\ %\hline
HETER~\cite{MillerHG17}  & Humor      & 1,780        & 71.4\%   & 5K    \\ %\hline        
TV~\cite{Boyd-GraberGZ13asist} & Spoiler  & 13,447   & 52.5\%   & 20K     \\ %\hline
\rowcolor{Gray}
BOOK~\cite{WanMNM19acl}    & Spoiler    & 17.67M & 3.2\%      & 373K \\ %\hline
EVAL~\cite{HuaNBW19naacl}   & Argument   & 10,386    & 38.3\%  & 8K     \\ %\hline
REQ~\cite{HuaNBW19naacl}    & Argument  & 10,386    & 18.4\%    & 8K   \\ %\hline
FACT~\cite{HuaNBW19naacl}   & Argument  & 10,386    & 36.5\%    & 8K  \\ %\hline
REF~\cite{HuaNBW19naacl}   & Argument   & 10,386    & 2.0\%     & 8K    \\ %\hline
QUOTE~\cite{HuaNBW19naacl}   & Argument & 10,386    & 1.6\%      & 8K   \\ %\hline
ARGUE~\cite{StabMSRG18emnlp}  & Argument  & 23,450  & 43.7\%    & 21K   \\ %\hline
SUPPORT~\cite{StabMSRG18emnlp} & Argument  & 23,450  & 19.4\%      & 21K \\ %\hline
AGAINST~\cite{StabMSRG18emnlp}  & Argument & 23,450  & 24.3\%     & 21K  \\ %\hline
\rowcolor{Gray}
AMAZON~\cite{ZhangZL15nips} & Sentiment  & 3.6M & 50.0\%    & 1M   \\ %\hline
\rowcolor{Gray}
YELP~\cite{ZhangZL15nips}   & Sentiment   & 560,000  & 50.0\%  & 232K    \\\midrule
\multicolumn{5}{c}{\textbf{Additional Datasets}} \\\midrule
\rowcolor{Gray}
FUNNY* & Humor & 244,428  & 50.0\%  & 171K \\ %\hline
\rowcolor{Gray}
BOOK*  & Spoiler  & 1.14M & 50.0\% & 112K \\ \bottomrule
\end{tabular}
}
\vspace{-3mm}
\end{table}

%that are largely diversed in the 21 selective datasets. 
\section{Datasets}  \label{sec:dataset}

We curate a collection of 21 textual datasets to evaluate the quality of different models on the semantic tagging task. To the best of our knowledge, our datasets have the best coverage of the real-world datasets for semantic tagging. We focus on three characteristics of the datasets that are crucial to a model's performance: size, label ratio, and label cleanliness. Datasets in our collection have 1,700 to 17,000,000 examples. 
% The sizes of these datasets range from 1,700 to 17,000,000. In addition to the size, we include another two dataset characteristics
% that we found closely correlated to the models' performance: label skewness and label cleanliness. These two characteristics are largely diverse in our selected datasets. 
The ratio of positive instances in our datasets ranges from 1.6\% to 71.4\%.
%For example, the ratio of positive instances ranges from 1.6\% to 71.4\%. 
4 of the 21 datasets use labels generated based on incomplete metadata, while the rest use either human-annotated labels or labels which are derived from complete metadata.

\noindent\textbf{Source.} We collected our datasets from various semantic tagging applications that are surveyed in Section~\ref{sec:apps}. We describe the collection process for each dataset below.
 
\begin{itemize}[leftmargin=*,noitemsep,topsep=0pt]
\item \textbf{Tip:} We use the SUGG dataset from the 9th task of SEMEVAL competition 2019~\cite{semeval19NegiDB}. The labels are based on whether a user comment can help improve windows software. The HOTEL~\cite{NegiB15emnlp} dataset is based on hotel reviews. The labels are derived based on whether a review sentence provides suggestions for future customers. The SENT~\cite{WangPWZ19msr} and PARA~\cite{WangPWZ19msr} datasets also contain tip sentences (customers' recommendations for software updates) and non-tip sentences (e.g., customers' experience of using the software). 
% SENT/PARA is annotated in a way that each record is a sentence/paragraph.
 
\item \textbf{Humor:} 
%FUNNY, HOMO, and HETER are 3 datasets obtained from Humor. FUNNY comes from the 
We use the FUNNY~\cite{MoralesZ17emnlp} dataset from the YELP Dataset Challenge~\cite{yelp_challenge}, in which a review received votes from readers if they think this review is funny. Reviews with more than 5 votes are annotated as positive and reviews with 0 votes are annotated as negative~\cite{MoralesZ17emnlp}. We use the HOMO and HETER datasets from the 7th task of SEMEVAL competition 2019~\cite{MillerHG17}. A sentence is labeled positive or negative depending on the occurrence of pun words. %HOMO pun uses the same word to express two meanings, while HETER pun uses two different words, which have similar pronunciation, to express two meanings. 
 
\item \textbf{Spoiler:} The TV~\cite{Boyd-GraberGZ13asist} and BOOK~\cite{WanMNM19acl} datasets are obtained from TV show comments and book comments, respectively. Each sentence is labeled depending on whether the sentence contains a spoiler. Having more than 17 million examples, BOOK is the largest among all the datasets.
 
\item \textbf{Argument:} We have 8 datasets for Argument application namely, EVAL, REQ, FACT, REF, QUOTE, ARGUMENT, SUPPORT, and AGAINST. Each dataset has been derived from two multi-class tagging tasks. EVAL, REQ, FACT, REF, and QUOTE~\cite{HuaNBW19naacl} include examples of different types of sentences that make up a paper review. ARGUMENT, SUPPORT, and AGAINST~\cite{StabMSRG18emnlp} are obtained from online discussions of controversial topics and contain sentences corresponding to argument opinion, supportive opinion, and opposed opinion, respectively. 
 
\item \textbf{Sentiment:} AMAZON~\cite{ZhangZL15nips} and YELP~\cite{ZhangZL15nips} are two Sentiment datasets that are obtained from AMAZON and YELP reviews, respectively. A review is annotated as positive if the writer gives a rating of 3 or 4 and as negative if the rating is 1 or 2.  
\end{itemize}

FUNNY* and BOOK* are 2 additional large datasets that we obtained from FUNNY and BOOK by balancing the positive and negative labels. We randomly drop a number of negative labels to make the label ratio balanced. Overall, the datasets we collected offer a good representation of real-world datasets and enable us to understand the strengths and limitations of different models.\\

% statistic, task, scale, P/N
\noindent\textbf{Dataset preparation.} Since dataset preparation can directly influence tagging performance and model selection, it is a critical component of the tagging pipeline. Our datasets come from different papers and public competitions, and therefore, must be transformed into the same format.
% To have comprehensive coverage of dataset characteristics for our study, we have collected 17 datasets from five types of surveyed sentences. They come from either published papers or public competitions~\cite{semeval19NegiDB, MillerHG17}. 
%We transformed the collected datasets into the same format. 
A data record is in the format of (text, label), with label 1 representing the text we wish to tag and 0 otherwise. We calculate crucial information about these datasets in Table~\ref{table:datasets}, such as the application that a dataset belongs to, the number of records, and the percentage of positive labels. %Since only 2 (FUNNY and BOOK) out of the 17 datasets are large, we include additional 4 large datasets. Specifically, AMAZON and YELP come from sentiment analysis, i.e. predicting whether a review expresses a positive opinion or not; FUNNY* and BOOK* are derived from FUNNY and BOOK datasets while their positive and negative ratio is balanced (50\%:50\%). We do not include any dataset from Product Description as they are not publicly available.
\\

\noindent\textbf{Dataset Characteristics.} We next describe the characteristics of the various datasets we collected/generated.
% size, coverage
\begin{itemize}[leftmargin=*,noitemsep,topsep=0pt]
\item \textbf{Size:} The scales of the datasets skew towards two ends. 15 out of 21 datasets have a small number of records at the scale of 1,000 to 99,999 (small datasets), while the remaining 6 datasets contain 100,000 to 20,000,000 records (large datasets). Small datasets are more common than large datasets in actual cases. This is because every record of a dataset requires human annotation. The high price of recruiting human annotators limits the scale of datasets. However, accompanying the emergence of experience-sharing websites such as Amazon and YELP, more and more large datasets appear. Following this, numerous customers share their opinions on these websites through sentence annotation. For example, YELP users share their opinions on YELP about whether a review is funny~\cite{yelp_challenge} or not. Given the polarization of real-word datasets in scale, a practical tagging model should be able to tag both small and large datasets accurately.  % a good solution should excel in both small datasets and large datasets as the are both common

%two datasets (i.e. HOMO and HETER) have more positive labels and the percentages are more than 70\%. The remaining 5 datasets TV, AMAZON, YELP, FUNNY*, and BOOK* exhibit a relatively balanced ratio of around 50\%. 
\item \textbf{Label Ratio:} Label imbalance is a common phenomenon in our datasets. We observed 14 out of 21 datasets having fewer positive labels than negative labels. More specifically, 10 of the 14 datasets exhibit a ratio of positive labels smaller than 25\% (considered as imbalance). For the remaining 7 datasets, 5 datasets exhibit a balanced ratio of around 50\% and 2 datasets exhibit a ratio of positive labels larger than 70\%. A higher percentage of positive labels is favorable due to two main reasons. First, more positive instances can increase the size of effective training data, leading to better training quality. Second, more positive instances can minimize the misleading effects of negative instances, increasing the accuracy of tagging results. 

\item \textbf{Cleanliness:} There are two common ways to obtain data labels: rule generation and human annotation. Relying on rules can potentially introduce dirty labels %in the datasets. 
due to missing annotations. 
For example, relying on number of votes to generate label FUNNY can introduce dirty labels for new records which do not have enough votes.
% For example, the generation of FUNNY that is based on the number of votes may meet such issue that some records are too new to accumulate enough votes. Therefore, these new records, which are potentially positive, may be mislabeled as negative that increase the dirtiness of datasets. 
Similarly, relying on spoiler alters to generate label BOOK can be dirty if writers do not leave any alert on their reviews.
% Similarly, the generation of BOOK that is based on the spoiler alerts may meet such issue that some writers do not leave any alert on their reviews, leading to mislabeling of potential positive records. 
Consequently, we consider FUNNY, BOOK, and their derivatives (FUNNY* and BOOK*), based on rule generation, as dirty datasets. The rest of the datasets, which are based on human annotation, are categorized as clean datasets. 
%\nikita{briefly mention if the human annotations are based on crowdworkers or experts. If they are based on crowdsourcing, we might have to be careful in defending that.} \jinfeng{The cause of dirtiness here is missing annotations. Some records lack acannotations such as votes or spoiler alerts. I use “missing annotations” instead of “noisy annotations” to make the concept of “dirty” more clear. Thanks for pointing out this point.}
\end{itemize}

%% file: experiment.tex
\newcommand{\largeh}{\ensuremath{\mathsf{Large\text{-}H}}}
\newcommand{\smallh}{\ensuremath{\mathsf{Small\text{-}H}}}
\newcommand{\largel}{\ensuremath{\mathsf{Large\text{-}L}}}
\newcommand{\smalll}{\ensuremath{\mathsf{Small\text{-}L}}}

% \begin{table}[]
% \centering
% \scalebox{0.75}{
% \begin{tabular}{|l|l|l|l|l|l|}
% \hline
% Dataset & \#Record     & \#Positive & \#Negative & \%Positive & App        \\ 
% \hline
% AMAZON	&3,600,000	&1,800,000	&1,800,000	&50.0\%	&Sentiment \\
% \hline
% YELP	&560,000	&280,000	&280,000	&50.0\%	&Sentiment \\
% \hline
% FUNNY* & 244,428   & 122,214 & 122,214 & 50.0\% & Humor   \\ \hline
% BOOK*  & 1,139,448 & 569,724 & 569,724 & 50.0\% & Spoiler \\ \hline
% \end{tabular}
% }
% \caption{Additional datasets}
% \label{table:additional_datasets}
% \end{table}

%\section{Evaluation}
%\section{Experiments}
\vspace{-1mm}
\section{Empirical Evaluation}
\label{sec:experiment}
% In this section, we compare the tagging performance of representative models, including LR, SVM, CNN, LSTM, and BERT on the datasets we collected. 
% % . We use 21 datasets with various characteristics as the tagging targets to construct the comprehensive comparison. 
% Our comparison results show that deep models are not necessarily better than simple models. Briefly, BERT indeed achieves higher F1 scores on small datasets. However, on large datasets, its F1 scores are not always higher than simple models. 

In this section, we compare the tagging performance of representative models, including LR, SVM, CNN, LSTM, and BERT on the datasets we collected. Experiments show that deep models do not consistently outperform the simple models across all datasets. On large datasets, in particular, deep models obtain similar or worse performance, yet they take significantly more training time. 

The section is organized as follows. We introduce experimental settings in Section~\ref{setting}. We report detailed performance of different models in Section~\ref{sec:detailed_analysis}. We focus on analyzing BERT in Section~\ref{subsec:representative} since BERT consistently outperforms other deep models.

\subsection{Experimental Settings} \label{setting}

\setlength{\tabcolsep}{5pt}
\begin{table}[t]
\caption{Dataset taxonomy}
\label{table:category}
\centering
\scalebox{0.9}{
\begin{tabular}{cl}
\toprule
Category & Datasets \\ \midrule
\multirow{2}{*}{\smalll} & HOTEL, SENT, PARA, REQ, \\
 & REF, QUOTE, SUPPORT, AGAINST \\ \midrule
\smallh & SUGG, HOMO, HETER, TV, EVAL, FACT, ARGUE             \\ \midrule
\largel & FUNNY, BOOK                                          \\ \midrule
\largeh & AMAZON, YELP, FUNNY*, BOOK*                          \\ \bottomrule
\end{tabular}
}
\vspace{-3mm}
\end{table}

\noindent\textbf{Dataset taxonomy.} We had introduced the 21 datasets in Section~\ref{sec:dataset}. We categorize the datasets into four groups based on size and label ratio (whether the number of records is larger than 100,000 and the percentage of positive labels is larger than 25\%). The dataset taxonomy is shown in Table~\ref{table:category}. It is based on 4 categories namely, small size low percentage (\smalll), small size high percentage (\smallh), large size low percentage (\largel), and large size high percentage (\largeh.). The taxonomy helps understand models' performance in terms of characteristics of the datasets.\\
% To better understand the performance difference, we divide the 21 datasets into four categories, according to the size and label ratio (whether the number of records is larger than 100,000 and the percentage of positive labels is larger than 25\%).
% We listed the dataset taxonomy in Table~\ref{table:category}. 
% Specifically, we divided these datasets into 4 groups, which are small size low percentage (\smalll), small size high percentage (\smallh), large size low percentage (\largel), and large size high percentage (\largeh.). 
% This taxonomy helps us evaluate whether deep models outperform simple models under different scenarios. \\

%\yuliang{Need to explain why we are dividing this way.} \jinfeng{Added one sentence to the end of the paragraph}
%\yuliang{modified the first sentence. Pls check.}

% environment

\noindent\textbf{Dataset Preparation.}  We split each dataset into a train set and a test set. Train set contains 80\% records while test set contains the remaining 20\% records. We do not split SUGG, HOMO, and HETER, since each of them already contains a separate test set for SEMEVAL competitions~\cite{semeval19NegiDB, MillerHG17}. The train set ratio of SUGG, HOMO, and HETER is 93\%, 80\%, and 80\%, respectively. \\

\noindent\textbf{Computing resource.} We conduct experiments on an AMAZON EC2 p3.8xlarge GPU server. Our instance is equipped with 4 Tesla V100 GPUs. Each GPU has 16GB memory, 640 Tensor cores, and 5120 CUDA cores. The server is running on Linux Ubuntu 16.04. 
The monetary cost is \$3 US dollars per hour of using the GPUs. \\%That is, the efficiency of a model matters because long running jobs could cost a lot. \\

% model tuning
\noindent\textbf{Hyper-parameters.}
We tune each model to the best performance according to common practices. We found using a combination of unigram and bigram in Bag-of-words representation yields the best tagging quality for LR and SVM. We adopt the default setting for BERT~\cite{devlin2018bert}. We set the batch size to 32, the max sequence length to 128, and the number of epochs to 3. For CNN and LSTM, we use the same batch size as BERT but set the max sequence length to 256 and the number of epochs to 10. We do not observe clear performance improvement when using larger batch size, longer max sequence length, and more epochs for CNN and LSTM. \\
%\yuliang{also mentioned how we implement those models. For example, LR and SVM are from sklearn. CNN and LSTM are re-implementation using pytorch. BERT uses huggingface. etc.} \jinfeng{we will make the code open source} \\

\noindent\textbf{Evaluation Metric.} We measure the quality as well as the training cost of tagging positive instances. 
We use F1 and training time accordingly. F1 is an averaged quality indicator and defined as 
$\frac{2 \cdot \mathsf{precision} \cdot \mathsf{recall}}{\mathsf{precision} + \mathsf{recall}}$, 
where precision and recall are standard information retrieval metrics. For example, if we assume that there are 10 positive instances and an algorithm tags 8 positive instances with 6 are correct, the precision is $6/8 = 0.75$ and recall is $6/10 = 0.6$. 
In this case, the F1 equals to $ 2 * 0.75 * 0.6 / (0.75 + 0.6) = 0.66$.\\

\noindent\textbf{Macro- and Micro-average F1} When evaluating a set of datasets, we report macro- and micro-average F1 to compare the overall performance of simple and deep models. Macro F1 is the average F1 score, for which it is insensitive to the sizes of datasets. Given that the 21 datasets are of various sizes, we also calculate micro-average F1, the sum of weighted F1s, as a complement to macro-average F1. Specifically, the weight of a dataset is the number of records of this dataset divided by total number of records of all datasets. Therefore, a larger dataset will have a higher weight than a smaller dataset. 

%A larger dataset has a higher weight, thus can reflect more of its F1 to the micro-average F1 score.} 

%$precision = \frac{\#\ true\ positive\ prediction}{\#\ positive\ predication}$ and $recall=\frac{\#\ true\ positive\ prediction}{\#\ true\ positive}$.
%$precision = positive prediction $ of prevision and recall we used F1 by default. \xiaolan{Do we need to add in-line equations for F1?  $F1 = \frac{2\cdot precision \cdot recall}{precision + recall}$} 

%We also report Accuracy or AUC as needed when comparing with the best quality known so far \xiaolan{I don't understand the previous sentence.}. For efficiency measurement, we evaluate training time, prediction time and model size.

\begin{figure*}[t]
    \centering
    \scalebox{0.8}{
    \includegraphics[width=\linewidth]{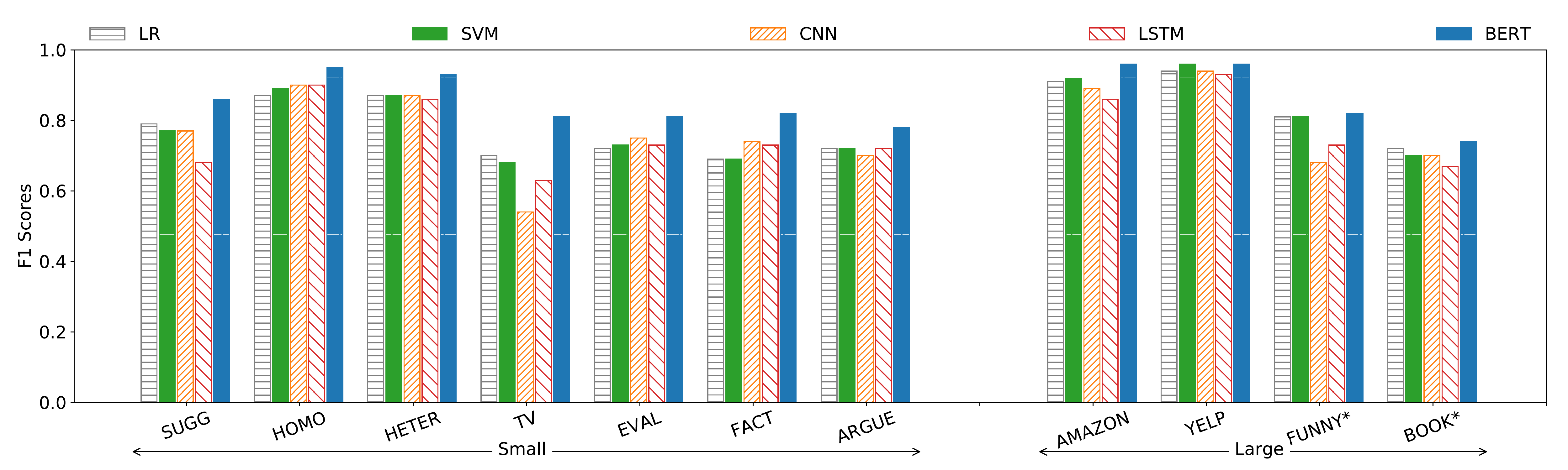}
    }
    \caption{F1 on \smallh\ and \largeh\ datasets with $\geqslant$ 25\% positive labels}
    \label{fig:f1_high_p}
\end{figure*}
\begin{figure*}[t]
    \centering
    \scalebox{0.8}{
    \includegraphics[width=\linewidth]{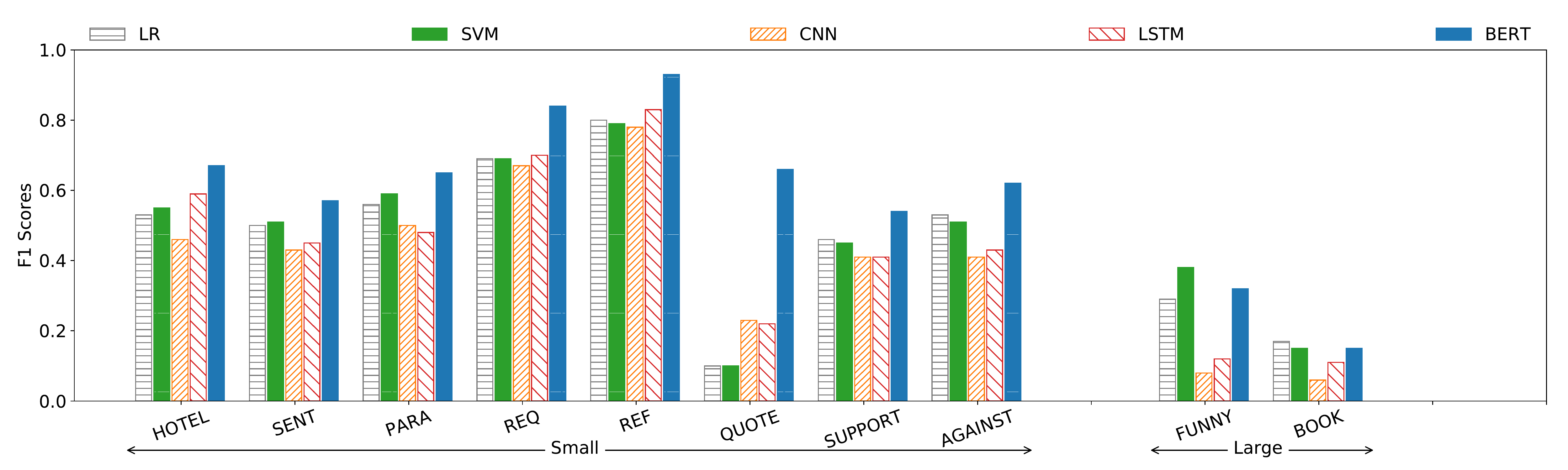}
    }
    \caption{F1 on \smalll\ and \largel\ datasets with $<$ 25\% positive labels}
    \label{fig:f1_low_p}
\end{figure*}

% \subsection{Detailed Performance}
\subsection{Result Analysis}
\label{sec:detailed_analysis}

% To comprehensively evaluate whether deep models indeed outperform simple models, we compare tagging F1s of LR, SVM, CNN, LSTM, and BERT on all 21 datasets. 

%We first compare how different algorithms perform on individual datasets. 

We first describe the performance of each individual model. Next, we compare the performance of different models on different dataset categories. Lastly, we discuss their performance and training time trade-offs.

% \subsubsection{F1 of different algorithms}
%\subsubsection{Performance of different models}
% \subsubsection{Comparison on different datasets}
\subsubsection{Performance of individual models}
% We plot F1 histograms with regard to different types of datasets for better comparison. Figure~\ref{fig:f1_high_p} shows the performance of datasets, whose percentages of positive labels are larger than 25\%. Datasets with positive ratios smaller than 25\% are reported in Figure~\ref{fig:f1_low_p}. \\

Table~\ref{table:avg_f1_category} shows the macro- and micro- average F1 score of each individual model on datasets grouped according to the taxonomy. Figure~\ref{fig:f1_high_p} and Figure~\ref{fig:f1_low_p} show the F1 scores on each dataset. Figure~\ref{fig:f1_high_p} shows the
F1 scores of individual model on the small and large datasets with high positive label ratio ($\geqslant$ 25\%). On the other hand, Figure~\ref{fig:f1_low_p} shows the F1 scores on the small and large datasets with low positive label ratio (i.e. imbalance). \\

% We present F1 scores of individual model according to dataset taxonomy in Table~\ref{table:category} with Figure~\ref{fig:f1_high_p} and Figure~\ref{fig:f1_low_p}. Figure~\ref{fig:f1_high_p} shows the
% F1 scores of different algorithms on the small and large datasets with high positive label ratio ($\geqslant$ 25\%). On the other hand, Figure~\ref{fig:f1_low_p} shows the F1 scores on the small and large datasets with low positive label ratio. \\

% Performance on different algorithms (LR, SVM, CNN, LSTM, BERT)
\noindent\textbf{LR and SVM.} 
LR and SVM achieve very similar F1 (difference $<\pm 0.03$) on all the datasets except FUNNY. Both LR and SVM achieve the lowest F1 (0.10 and 0.10) on QUOTE and the highest F1 (0.94 and 0.96) on YELP. Compared with QUOTE, YELP has a larger size (560,000 vs. 10,000) and a higher label ratio (0.5 vs. 0.02), indicating the size or the label ratio of a dataset has significant effect on the performance of simple models. As shown by Figure~\ref{fig:f1_high_p} and Figure~\ref{fig:f1_low_p}, LR achieves an average F1 of 0.79 on the datasets with the positive label ratio  $\geqslant25\%$ and 0.46 on the positive label ratio $<25\%$. SVM shows a similar behavior on datasets with different ratios of positive labels. This suggests that the tagging quality of simple models is significantly affected by the ratio of positive labels. \\

\noindent\textbf{CNN and LSTM.} 
The F1 score of CNN ranges from 0.08 to 0.94, while the F1 score of LSTM ranges from 0.11 to 0.93. Interestingly, these scores are lower than F1 scores of LR and SVM, indicating that deep models do not always outperform simple models. Our finding contradicts the popular belief that deep models are the best choice for semantic tagging task. Furthermore, we find that CNN and LSTM are also sensitive to positive label ratio. The average F1 of CNN is 0.77 on datasets with a ratio $\geqslant$ 25\%, and 0.39 on datasets with a ratio $<$ 25\%.\\
%, as shown in Figure~\ref{fig:f1_high_p} and Figure~\ref{fig:f1_low_p}, respectively. \\

\noindent\textbf{BERT.} 
% As a representative deep model, BERT is more in line with people's expectations that deep models outperform simple models. 
BERT gains wide-spread popularity in recent years and is considered a de-facto model for semantic tagging tasks. Not surprisingly, it does achieve highest F1 scores on most of the datasets (19 of 21). This is perhaps because it is pre-trained on a large corpus and hence is optimized for large datasets. However, our experiments reveal that BERT does not show apparent advantages over simple models on large datasets. The average F1 scores of BERT, LR, and SVM on all the 6 large datasets are 0.66, 0.64, 0.66, respectively. In fact, on 2 of the large datasets which are also imbalanced, BERT performs worse than simple models. These results suggest that deep models may not always perform better, especially on large or imbalanced datasets. \\

\begin{figure}[!ht]
    \centering
    \begin{subfigure}{0.52\linewidth}	    \includegraphics[width=\linewidth]{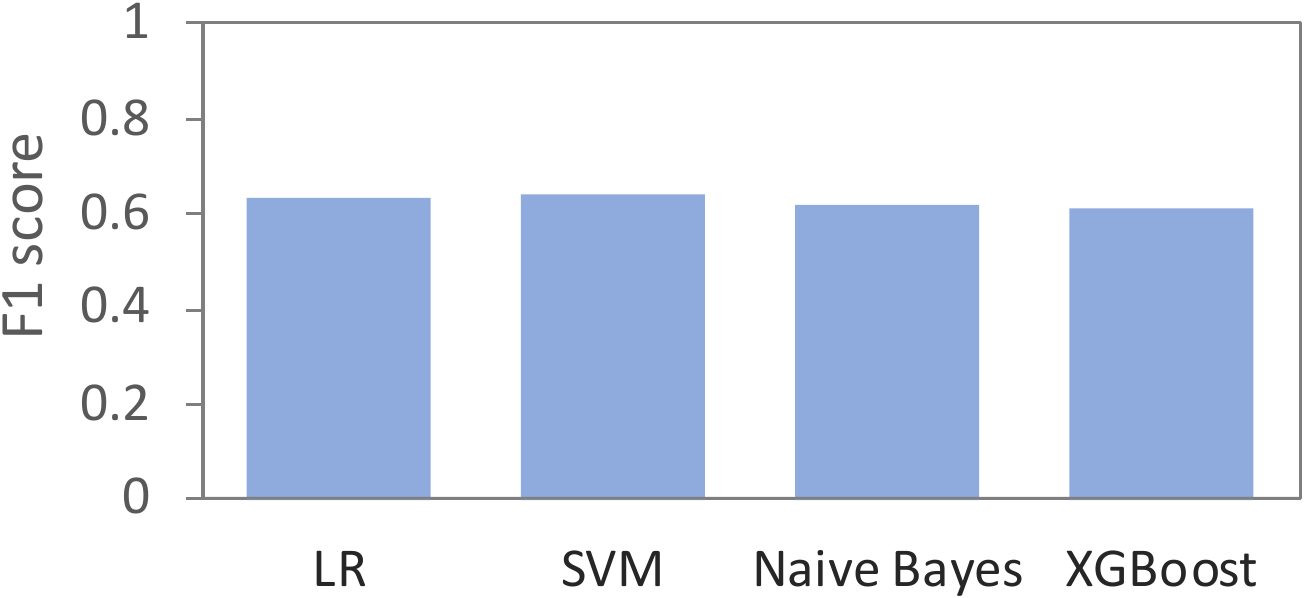}
	    \caption{Simple models}
	    \label{fig:naive_bayes_xgb_avg-crop.pdf}				
	\end{subfigure}	
    \begin{subfigure}{0.47\linewidth}
	    \includegraphics[width=\linewidth]{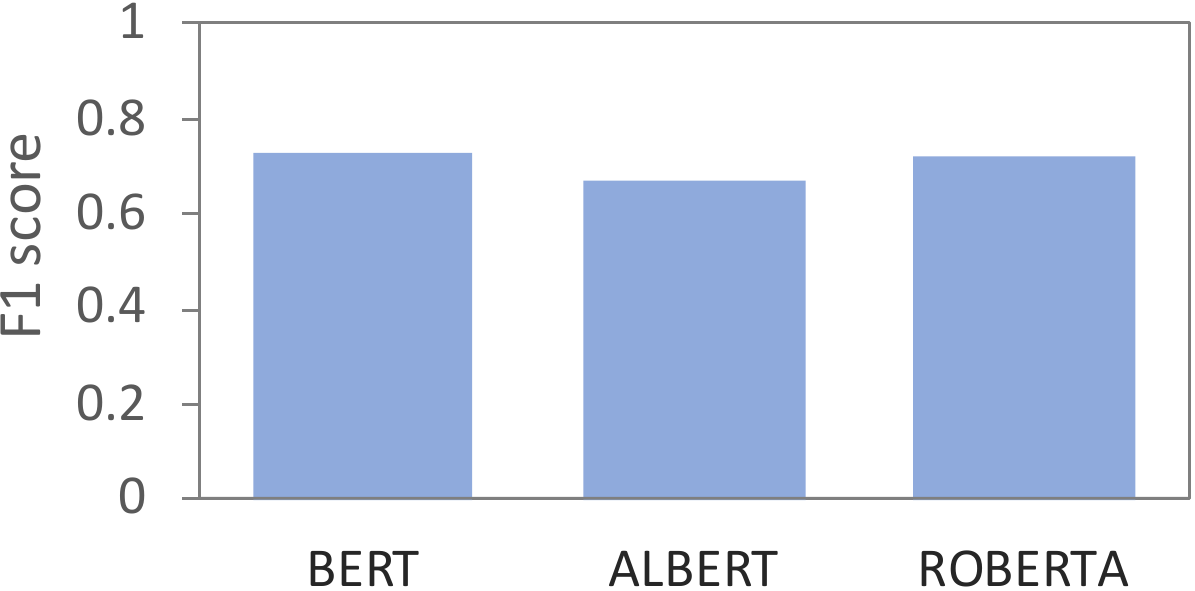}
	    \caption{Deep models}
	    \label{fig:albert_roberta_avg-crop.pdf}	
	\end{subfigure}	    
    \caption{The (macro) average F1 of industrial models on all 21 datasets}
    \label{fig:industrial_models}
    \vspace{-2mm}
\end{figure}

\noindent\textbf{Other industrial models.} We also evaluate the performance of more simple and deep models that are extensively used in industry. The newly investigated simple models include Naive Bayes~\cite{naive_bayesian} and XGBoost~\cite{ChenG16kdd} (a ensemble/boosting model). The newly investigated deep models include ALBERT~\cite{lan2019albert} and ROBERTA~\cite{roberta}, which are obtained from Huggingface transformers~\cite{huggingface}. We report average F1 scores of Naive Bayes and XGBoost in Figure~\ref{fig:naive_bayes_xgb_avg-crop.pdf}. 
We also include the results of LR and SVM in the same figure for comparison. The average F1 of LR, SVM, Naive Bayes, and XGBoost are overall similar, so we use LR and SVM as representative simple models in our paper. We report average F1 of ALBERT and ROBERTA in Figure~\ref{fig:albert_roberta_avg-crop.pdf}. We also include F1 of BERT in the same figure for comparison. Overall, BERT shows slightly better F1 than ALBERT and ROBERTA, so we take BERT as the representative attention-based deep model in our paper. We report the results of F1 scores for individual datasets in Appendix of the technical report~\cite{technical_report}. %\xiaolan{Maybe consider replacing figure 3 with a table. If keep using the bar graph, consider using different colors to represent the model as other figures do. Also, maybe consider highlighting the key insights (newly added simple models perform similarly as LR/SVM, newly added deep models are not as good as BERT).   }

\begin{table}[t]
\caption{The (macro-/micro-) average F1 of LR, SVM, CNN, LSTM, and BERT in dataset categories}
\label{table:avg_f1_category}
\centering
\scalebox{0.9}{
\begin{tabular}{c|ccccc}
\toprule
 & LR & SVM & CNN & LSTM & BERT \\ \midrule
\largeh & 0.85/0.77 & 0.85/0.76 & 0.80/0.72 & 0.80/0.72 & 0.87/0.79 \\ \midrule 
\smallh & 0.77/0.73 & 0.76/0.72 & 0.75/0.70 & 0.75/0.71 & 0.85/0.82 \\ \midrule
\smalll & 0.52/0.51 & 0.52/0.51 & 0.49/0.47 & 0.51/0.49 & 0.68/0.66 \\ \midrule
\largel & 0.23/0.20 & 0.27/0.20 & 0.07/0.06 & 0.12/0.11 & 0.24/0.19 \\ \bottomrule
\end{tabular}
}
\end{table}

% \subsubsection{F1 on different datasets} 
\subsubsection{Comparison on different dataset categories}
% \subsubsection{Performance on different dataset categories}
% We continue to analyze the performance of five models to understand what kinds datasets are favored by them. 
Next, we analyze the performance of the models by dataset categories. Table~\ref{table:avg_f1_category} shows the macro- and micro- average F1 scores of different models.

% large high
\smallskip
\noindent\textbf{Large and high datasets.} 
Overall, all models perform the best on datasets that are large in size and high in the ratio of positive labels. This is reflected in higher F1 of every model on the \largeh{} dataset category than on other categories. \\

%The average F1s of LR, SVM, CNN, LSTM, and BERT are 0.85, 0.85, 0.80, 0.80, and 0.87, respectively, that are higher than their average F1s on other three types of datasets. This suggests that both deep and simple tagging models favor \largeh\ datasets. \\ 

% small high 
\noindent\textbf{Small and high datasets.} 
The average F1 scores of different models on the \smallh{} datasets are slightly lower than F1 scores on \largeh{} datasets. On the other hand, average F1 scores on \smalll{} datasets are much lower. This suggests that the size of a dataset does not affect tagging quality significantly when the ratio of positive label is high. \\

% The average F1 scores of LR, SVM, CNN, LSTM, and BERT are 0.77, 0.76, 0.75, 0.75, and 0.85, respectively. These are slightly lower than those achieved by corresponding models on \largeh\ datasets. The small differences in F1 scores indicate that the size of a dataset does not affect tagging quality significantly when the ratio of positive label is high.  \\

% small low
\noindent\textbf{Small and low datasets.}
The average F1 scores on \smalll{} datasets are significantly lower than on \smallh{} datasets. 
% The average F1 scores of LR, SVM, CNN, LSTM, and BERT are 0.52, 0.52, 0.49, 0.51, and 0.68, respectively. These are significantly lower than those achieved by corresponding models on \smallh\ datasets. 
This indicates that a low ratio of positive labels (i.e. imbalance) can negatively affect the tagging quality of models. \\ %Specifically, simple models obtain very low F1 scores on QUOTE, SUPPORT, and AGAINST datasets from the Argument applications, indicating simple models may not be suitable for tagging Argument datasets.  \\

% large low
\noindent\textbf{Large and low datasets.}  
Intuitively, large number of labels offer more training examples which should lead to high F1. However, we find that F1 scores on \largel\ datasets do not follow this intuition. The average F1 scores of different models on \largel{} datasets are the lowest among the different dataset categories. There are two reasons for low F1 scores. First, ratios of positive labels of the two datasets are significantly lower compared with other datasets. Second, labels of the two datasets are dirtier than other datasets, as we mention in Section~\ref{sec:dataset}. Interestingly, simple models achieve even better F1s than deep models on the two datasets that may due to poor label cleanliness and/or severe imbalance. This suggests that model selection should be done more carefully for \largel\ datasets with many dirty labels. \\ 

\noindent\textbf{Multi-label datasets.} We also perform comparison on multi-label datasets. We find two datasets, BIO and DEF, from task 6 of the SEMEVAL2020 competition~\cite{semeval2020}. BIO is a Named Entity Recognition dataset while DEF is a Knowledge Graph Extraction dataset. BIO contains around 470, 000 labels so the dataset is large. Each label is associated with a word token, and is either B, I or O.  The best performing simple and deep models show similar F1s (same regarding labels B and O, 0.02 difference regarding label I). The results indicate that simple models can achieve similar performance as deep models when the number of labels is sufficiently large. DEF contains around 18, 000 labels so the dataset is small. Each label is associated with a sentence, and is either T or F. The best performing deep model outperforms the best simple model on F1 (0.14 gap regarding label T and 0.04 improvement regarding label F). The results indicate that deep models achieve better performance than simple models when the number of labels is small. We present the detailed results and discussions in Appendix of the technical report~\cite{technical_report}.

\begin{figure}[!ht]
    \centering
    \begin{subfigure}{0.46\linewidth}
	    \includegraphics[width=\linewidth]{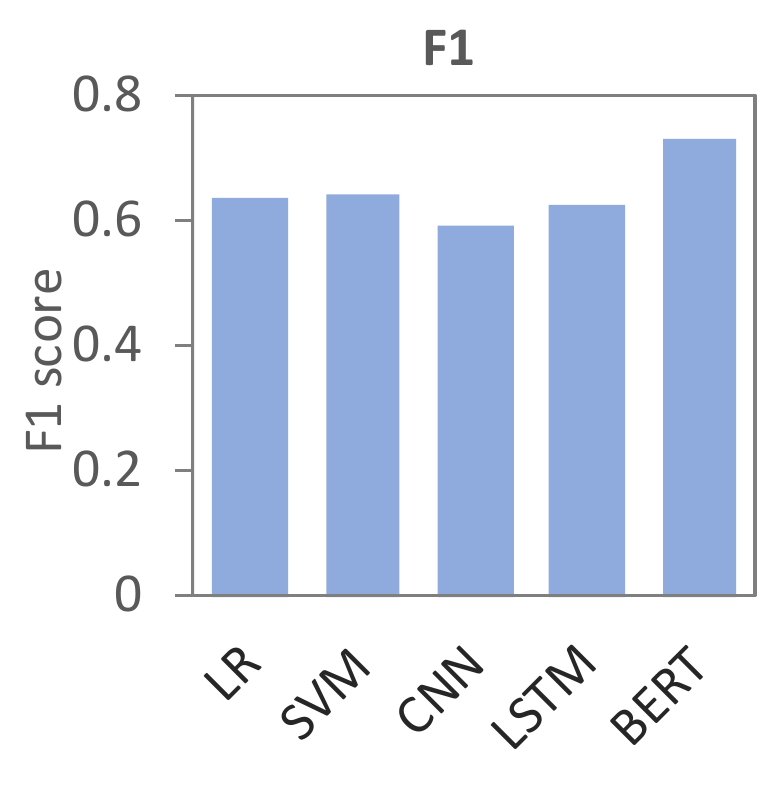}
	    \caption{F1}
	    \label{fig:avg_f1}				
	\end{subfigure}	
    \begin{subfigure}{0.49\linewidth}
	    \includegraphics[width=\linewidth]{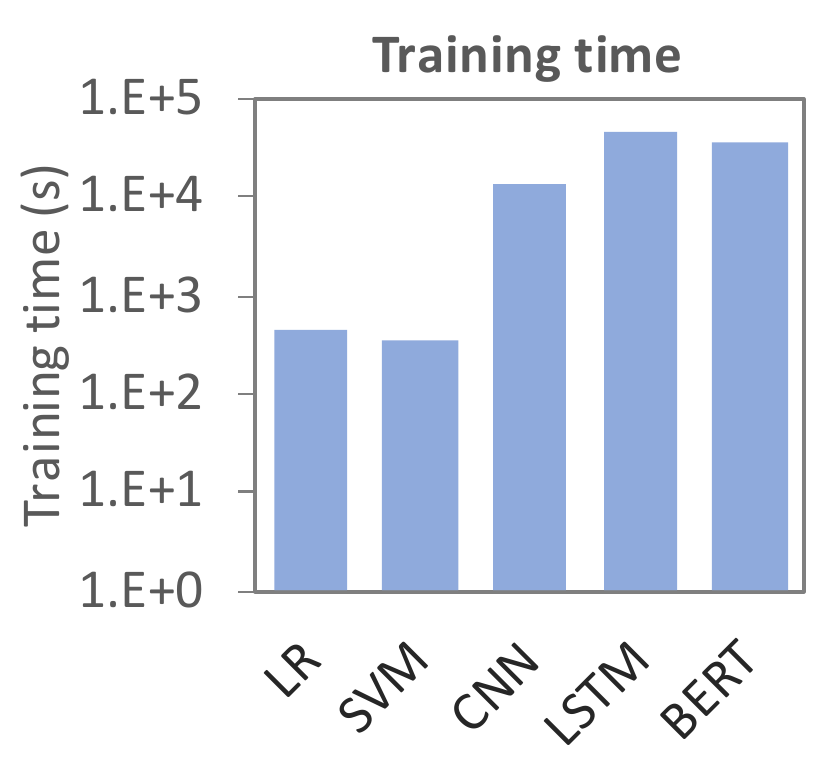}
	    \caption{Time (log-scaled)  }
	    \label{fig:avg_traintime}				
	\end{subfigure}	    
    \caption{The (macro) average F1 score and average training time (log-scaled) on all datasets}
    \label{fig:avg_f1_time}
    \vspace{-3mm}
\end{figure}

% \begin{figure}[t]
%     \centering
%     \scalebox{0.8}{
%     \includegraphics[width=\linewidth]{figure/exp/avg_f1.pdf}
%     }
%     \caption{Average F1 on all datasets}
%     \label{fig:avg_f1}
% \end{figure}

% \begin{figure}[t]
%     \centering
%     \scalebox{0.8}{
%     \includegraphics[width=\linewidth]{figure/exp/avg_traintime.pdf}
%     }
%     \caption{Average training time (in logarithmic seconds) on all datasets}
%     \label{fig:avg_traintime}
% \end{figure}

% \subsubsection{Average Performance} 
% \subsubsection{Overall comparison.} 
\subsubsection{Performance/Training time trade-off}
%We compared average F1 and training time of LR, SVM, CNN, LSTM, and BERT on all datasets. Since BERT achieves the overall best F1s, we also compared its tagging quality with the state-of-the-art results on individual datasets. 

Figure~\ref{fig:avg_f1} shows the (macro) average F1 scores of different models across all the datasets. We find that not all deep models outperform simple models. While BERT achieves higher F1 (with a large margin of 0.11) than LR/SVM, CNN and LSTM perform much worse.

We report the training time of the models in Figure~\ref{fig:avg_traintime}. We would like to remind the readers that all the deep models were trained on GPU while the simple models were trained using a CPU. Although some deep models achieve higher F1 than simple models, they also take 30x-130x more time for training. LSTM is the slowest and takes 13 hours on average for training and costs 39 dollars. 
% Given this is the price for a single experiment, multi-experiments are expected to be more expensive. 
CNN and BERT are slightly faster but still take a lot of time. In practice, debugging and parameter tuning deep models also take up a lot of time, adding to the overall cost. For example, BERT has a number of tunable hyperparameters, such as maximum sequence length, training batch size, number of training epochs, and learning rate. Each hyper-parameter has multiple options. Even with only 3 options per hyper-parameter, there can be as many as $3^4=81$ combinations that will cause significant overhead. On the other hand, simple models take less than 500 seconds using a single CPU. This suggests that simple models can bring more economic benefits to users, especially to those who have no access to GPU's but need to train tagging models on large datasets. \\
%\yuliang{changed the wording a little bit. Need to come back to this paragraph.}\jinfeng{updated the wording further} \\

% \begin{figure*}[!ht]
%     \centering
%     \scalebox{0.8}{
%     \includegraphics[width=\linewidth]{figure/exp/f1_delta.pdf}
%     }
%     \caption{F1 gain of BERT over the best of LR and SVM}
%     \label{fig:f1_delta}
% \end{figure*}

\vspace{-2mm}
\subsection{Analysis of BERT} \label{subsec:representative}
% highlighted Datasets: HOTEL, FUNNY ()

As shown in Figure~\ref{fig:f1_high_p} and Figure~\ref{fig:f1_low_p}, BERT consistently achieves higher F1 scores than other deep models. We, therefore, take a closer look at BERT.\\

% \noindent\textbf{BERT (the best quality).} As shown in Figure~\ref{fig:f1_high_p} and Figure~\ref{fig:f1_low_p}, BERT achieves the highest F1s on 19 out of 21 datasets. BERT does not outperform LR on BOOK, and also not win SVM on FUNNY, but their F1s are pretty close. This result shows that BERT is generally an accurate model across different domains and semantic tagging tasks. \\

\begin{figure}[t]
    \centering
    \includegraphics[width=\linewidth]{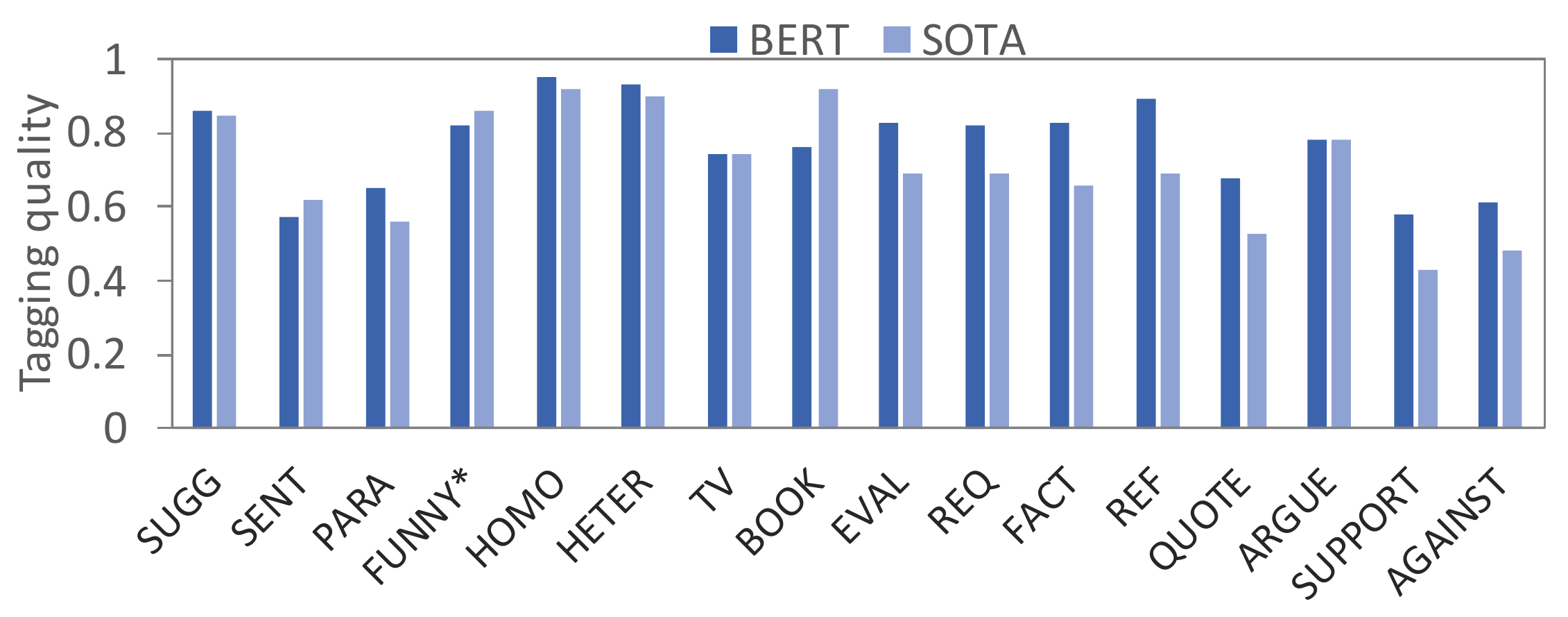}
    \caption{BERT compares with SOTA (the state-of-the-art). We follow the quality metric in SOTA publications (F1 by default, Accuracy for FUNNY \protect\cite{MoralesZ17emnlp}, TV \protect\cite{WanMNM19acl}, and AUC for BOOK \protect\cite{WanMNM19acl}.)}
    %\caption{BERT compares with SOTA (the state-of-the-art). We follow the quality metric in SOTA publications (F1 by default, Accuracy for FUNNY \protect\cite{MoralesZ17emnlp}, TV, and AUC for BOOK.)}
    \label{fig:bert_vs_sota}
    \vspace{-3mm}
\end{figure}

\noindent\textbf{BERT versus domain SOTA.} Our results indicate BERT as the most accurate generic model. We, therefore, compare it with the state-of-the-art (SOTA) models, which are considered as golden standards for semantic tagging tasks, on individual datasets. 
These SOTA results (e.g., 0.85 F1 on SUGG from SEMEVAL 2019 champion~\cite{semeval19liuWY}) were obtained by leveraging domain-specific knowledge.
% \yuliang{The following needs to be changed: (1) we directly took their numbers and (2) the use a table to summarize their techniques }\jinfeng{Emphasized (1) at the beginning of the next paragraph. Summarized (2) with one sentence above}
We follow the experimental settings and metrics given in published materials. 
Specifically, we compute the F1 score for SUGG~\cite{semeval19liuWY}, SENT~\cite{WangPWZ19msr}, PARA~\cite{WangPWZ19msr}, HOMO~\cite{ZouL19naacl}, HETER~\cite{DiaoL0FWZX19www}, EVAL~\cite{HuaNBW19naacl}, FACT~\cite{HuaNBW19naacl}, REF~\cite{HuaNBW19naacl}, QUOTE~\cite{HuaNBW19naacl}, ARGUE~\cite{StabMSRG18emnlp}, SUPPORT~\cite{StabMSRG18emnlp}, AGAINST~\cite{StabMSRG18emnlp}, and Accuracy for FUNNY*~\cite{MoralesZ17emnlp}, TV~\cite{WanMNM19acl} and AUC~\cite{WanMNM19acl} for BOOK~\cite{WanMNM19acl}. % \yuliang{stopped here.} 
Following the computing process, we directly take the published SOTA results from related publications and perform comparison between SOTA and BERT. 

As shown in Figure~\ref{fig:bert_vs_sota}, BERT achieves comparable or even better results compared with SOTA methods. Based on these findings, practitioners should really consider using BERT as a base model to replace existing solutions or develop new pipelines for semantic tagging. We also notice that BERT does not outperform %domain taggings 
SOTA in SENT, FUNNY*, and BOOK. Regarding SENT~\cite{WangPWZ19msr}, we can only obtain partial training set %from the authors
so the F1 scores of BERT and SOTA are not comparable. For FUNNY*~\cite{MoralesZ17emnlp}, the gap is pretty small (0.04). For BOOK~\cite{WanMNM19acl}, domain tagging extensively leverages the names of characters to identify spoilers, while these names are out of the vocabulary of BERT. Except for these extreme cases, BERT is always the best tagging model on most datasets. \\
% Performance 

\begin{figure}[t]
    \vspace{-3mm}
    \centering
    \scalebox{0.7}{
    \includegraphics[width=\linewidth]{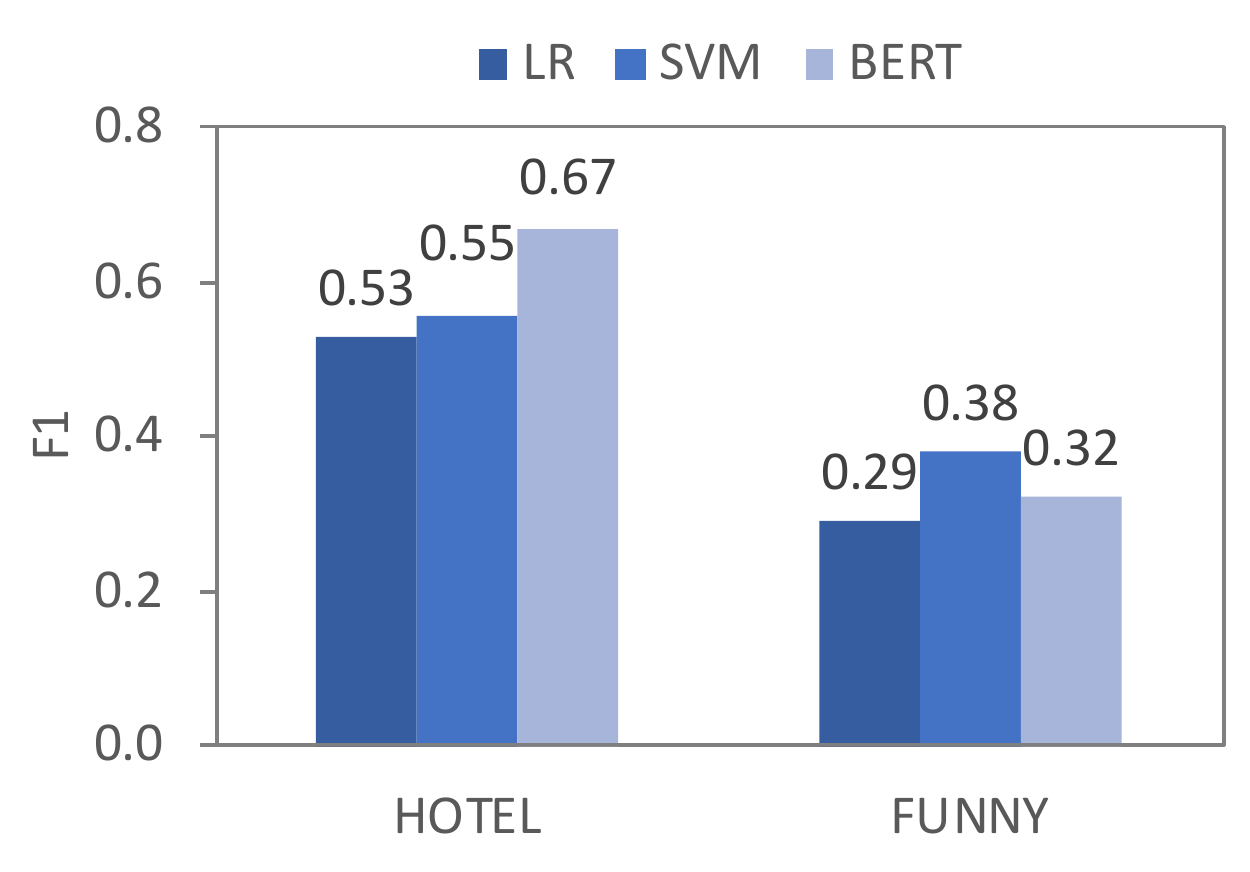}
    }
    \caption{F1s on a small/large dataset HOTEL/FUNNY}
    \label{fig:f1_representative}
    \vspace{-1mm}
\end{figure}

\noindent\textbf{BERT on representative datasets.}
% Since BERT achieves the higher F1s than CNN and LSTM on all 21 datasets, we choose BERT as representative deep model and compare it with simple models.
We choose HOTEL, FUNNY as the representative datasets for small, large datasets, respectively, and compare BERT with simple models. As shown in Figure~\ref{fig:f1_representative}, BERT achieves higher F1 than LR (0.14) and SVM (0.12) on HOTEL, confirming that BERT indeed significantly improves tagging F1 in some cases. However, in other cases, BERT does not outperform simple models. For example, BERT performs worse than SVM by 0.06 F1 on FUNNY. Simple models can sometimes outperform BERT while taking significantly less time for training.
% Not only that BERT is not necessarily better than simple models on F1, but also BERT takes significant training time when the dataset is large. 
For example, BERT takes 1.4 days to train FUNNY that contains 4.75 million records. Therefore, if considering both F1 improvement and time consumption as evaluation criteria, deep models are not the best choice for semantic tagging in some cases. \\

\begin{table}[!t]
\caption{F1 of LR and SVM without/with pre-trained BERT embeddings (eb.) }
\label{table:pre_train}
\centering
\scalebox{1}{
\begin{tabular}{c|cc|cc}
\toprule
Dataset & LR & LR + eb. & SVM & SVM + eb. \\
\midrule
HOMO & 0.87 & 0.94 & 0.89 & 0.93 \\
HETER & 0.87 & 0.92 & 0.87 & 0.91 \\
QUOTE & 0.10 & 0.35 & 0.10 & 0.34 \\
\bottomrule
\end{tabular}
}
\end{table}

\noindent\textbf{Effect of pre-trained embeddings} We conducted experiments to evaluate how much simple models can benefit from pre-trained embeddings. Given that BERT is similar to Word2Vec and Glove that use word vectors, for the convenience, we used BERT that was pre-trained over Wikipedia corpus as the embedder. For each input text, BERT outputs the last-layer [CLS] vector~\cite{devlin2018bert} as the featurization vector. We then ran LR and SVM on all featurization vectors and presented F1s of the most representative datasets (i.e. HOMO, HETER, and QUOTE) in Table~\ref{table:pre_train}. The results show that LR can achieve better F1s on small datasets with pre-trained embeddings. F1 improvements of HOMO, Heter, and QUOTE are 0.07, 0.05, and 0.25, respectively. Similarly, SVM also achieves better F1s on these three datasets. The F1 improvements of HOMO, HETER, and QUOTE are 0.04, 0.04, and 0.23, respectively. These results suggest that simple models can benefit from pre-trained embeddings. More results and discussions are in Appendix of the technical report~\cite{technical_report}.

% \subsection{Key Findings}
%\section{Key Results} 
\section{Analysis by dataset characteristics} \label{sec:effect_characteristic}
%\section{Comparison by dataset characteristics}
Our empirical evaluation suggests that model performance and, thus, model selection is influenced by dataset characteristics. We, therefore, investigate the effect of size, label ratio, and label cleanliness on model's performance. First, we compare BERT and simple models by dataset types to find more insights in Section~\ref{key_results}. Next, we investigate the effects of dataset characteristics on model selection in Section~\ref{sec:dataset_characteristic}. Lastly, we summarize our findings to help practitioners find the best model for their dataset in Section~\ref{subsec:heatmap}.

% Empirical evaluation shows that dataset characteristics are key factors to determine whether deep models outperform simple models. We, therefore, investigate how dataset characteristics impact the performance of deep and simple models. Specifically, we analyze the effect of size, label ratio, and cleanliness on models' performance. We found that simple models outperform deep models when datasets are large, fraction of positive examples is large or informative tokens are less clear.

% The rest of the section is organized as follows. We compare BERT and simple models by dataset types to find more insights in Section~\ref{key_results}. We investigate the effects of dataset characteristics on model selection in Section~\ref{sec:dataset_characteristic}. In the end, we give a summary on how our results guide practitioners to select the best model for their dataset in Section~\ref{subsec:heatmap}.

\vspace{2mm}
\subsection{Comparison of best models by type}\label{key_results}% summarization w.r.t. 
% summarize the following sections

\begin{table}[t]
\caption{Comparison of the best DEEP model and the best SIMPLE model on different types of datasets}
\label{table:best_dl_ml}
\centering
\begin{tabular}{c|ccc|cc}
\toprule
\multirow{2}{*}{Datasets} & \multicolumn{3}{c|}{Avg F1} & \multicolumn{2}{c}{Avg train time (s)} \\ % \cline{2-6} 
                              & DEEP      & SIMPLE      & $\Delta$   & DEEP                  & SIMPLE                \\ \midrule
\smalll                      & 0.68    & 0.52    & 0.16    & 308                 & $<$1                 \\ 
\smallh                      & 0.86    & 0.78    & 0.08    & 324                 & $<$1                 \\ \midrule
\largel                      & 0.24    & 0.27    & -0.03   & 308,680             & 3,128             \\ 
\largeh                      & 0.87    & 0.85    & 0.02    & 38,466           & 318             \\ \bottomrule
\end{tabular}
\vspace{-3mm}
\end{table}

In this section, we summarize our key findings. We only compare the best deep model (i.e. BERT, denoted as DEEP) and the best simple model (denoted as SIMPLE) for simplicity. We report the average F1 and average training time on each type of datasets in Table~\ref{table:best_dl_ml}.\\

\noindent\textbf{Model selection.} Our goal is to find out whether DEEP is better than SIMPLE. Our results show: (1). DEEP outperforms SIMPLE by a large margin on small datasets (0.16 on \smalll\ and 0.08 on \smallh\, respectively). However, it does not win on large datasets (-0.03 on \largel\ and 0.02 on \largeh). (2). Training DEEP is efficient on small datasets (in average 308/324 seconds for \smalll\ / \smallh), but inefficient on large datasets (in average 86/11 hours for \largel\ / \largeh). (3). SIMPLE becomes very competitive on large datasets since it obtains F1 similar to DEEP yet takes much less training time (on average 0.9/0.1 hours for \largel\ / \largeh). Therefore, DEEP will be a good choice for small datasets, but a questionable choice for large datasets.  \\%We compared the performance of individual algorithms in Section~\ref{sec:detailed_analysis} \\ 
% \yuliang{``runtime'' changes to ``training time''}\jinfeng{updated}

\noindent\textbf{Dataset preference.} We observed consistent dataset preference of DEEP and SIMPLE regarding F1. Specifically, (1). both DEEP and SIMPLE achieve the highest F1s on \largeh, the 2nd F1s on \smallh, the 3rd F1s on \smalll, and the smallest F1s on \largel. The performance of a model subjects to the dataset taxonomy. (2). Both DEEP and SIMPLE favor datasets with high percentages of positive labels. They achieve higher F1 on \smallh\ than \smalll\ (0.86 versus 0.68 for DEEP and 0.78 versus 0.52 for SIMPLE). (3). Neither DEEP nor SIMPLE achieves satisfactory performance on \largel. The F1s of DEEP and SIMPLE are 0.24 and 0.27, respectively, which are significantly smaller than F1s they obtained on other types of datasets, such as 0.68 and 0.52 on \smalll. 

\begin{figure}[!ht]
    \centering
    \begin{subfigure}{0.49\linewidth}
	    \includegraphics[width=\linewidth]{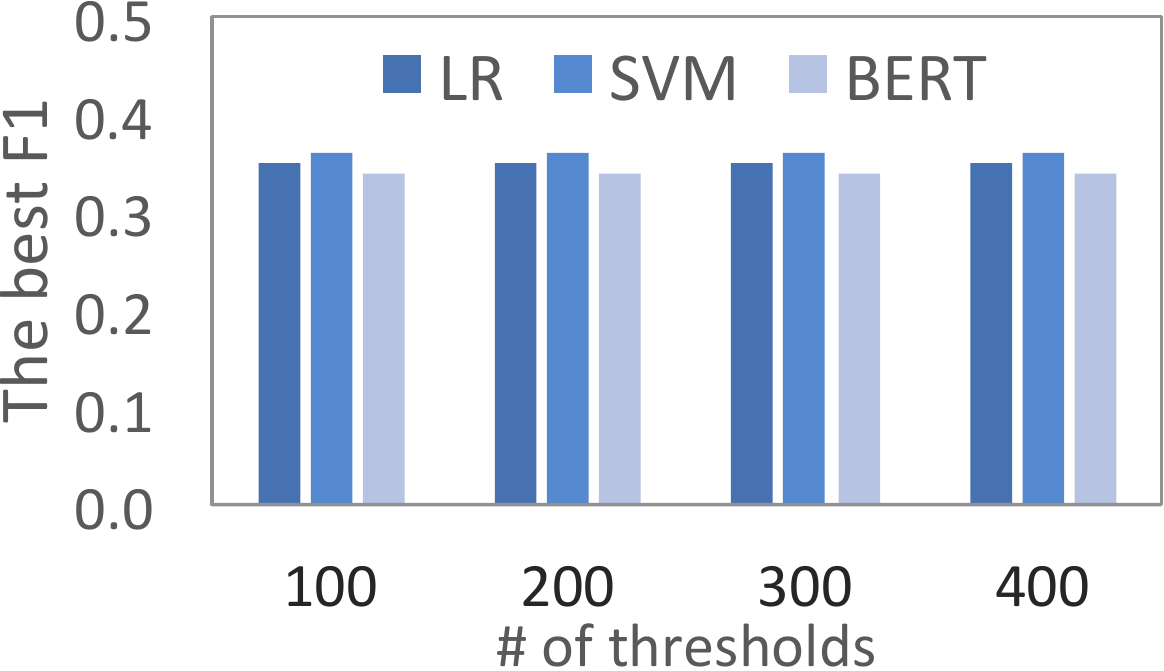}
	    \caption{FUNNY }
	    \label{fig:calibrate_funny}				
	\end{subfigure}	
    \begin{subfigure}{0.49\linewidth}
	    \includegraphics[width=\linewidth]{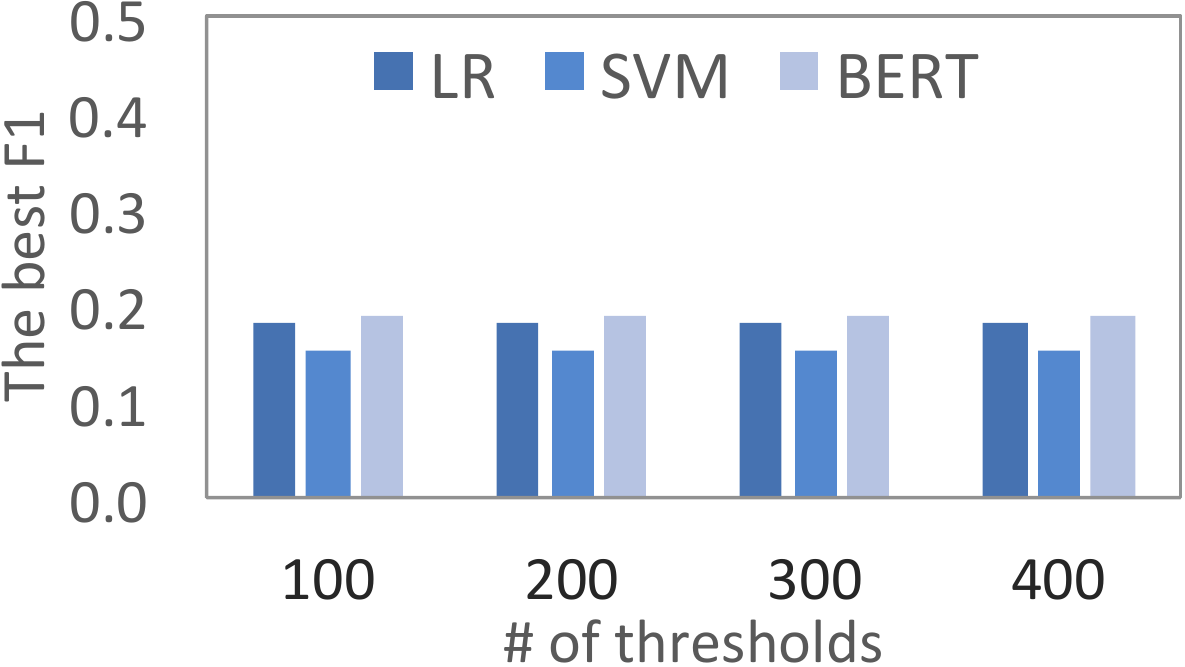}
	    \caption{BOOK}
	    \label{fig:calibrate_book}		
	\end{subfigure}	    
    \caption{The best F1 of LR, SVM, and BERT on FUNNY/BOOK (1 million labels) by varying classification thresholds}
    \label{fig:calibrate}
    \vspace{-3mm}
\end{figure}

\vspace{3mm}
\noindent\textbf{Large imbalanced datasets.} Our experiments show that SIMPLE achieves better F1 scores than DEEP on larger imbalanced datasets. Specifically, SVM outperforms BERT on FUNNY and LR outperforms BERT on BOOK. Considering that imbalance tackling techniques potentially affect the comparisons, we conducted new experiments to calibrate the classification threshold for each model (a popular technique). Figure~\ref{fig:calibrate} shows comparison results after calibration. The results show that simple models still show better F1s on FUNNY and similar F1s on BOOK, in comparison to BERT. Notably, We observe that adding calibration improves F1s of all models, especially BERT, but does not change the tendency that simple models perform similar or even better than BERT. More experiments are presented in Appendix of the technical report~\cite{technical_report}.

% \subsection{Further Analyses}
\vspace{2mm}
\subsection{Effect of dataset characteristics}\label{sec:dataset_characteristic}
In this section, we conduct analyses to understand the effects of dataset characteristics on the tagging quality. Our analyses consider the size of training set, the skewness of label ratio, and the existence of informative tokens. 

\begin{figure*}[!t]
    \centering
    \begin{subfigure}{0.24\linewidth}
	    \includegraphics[width=\linewidth]{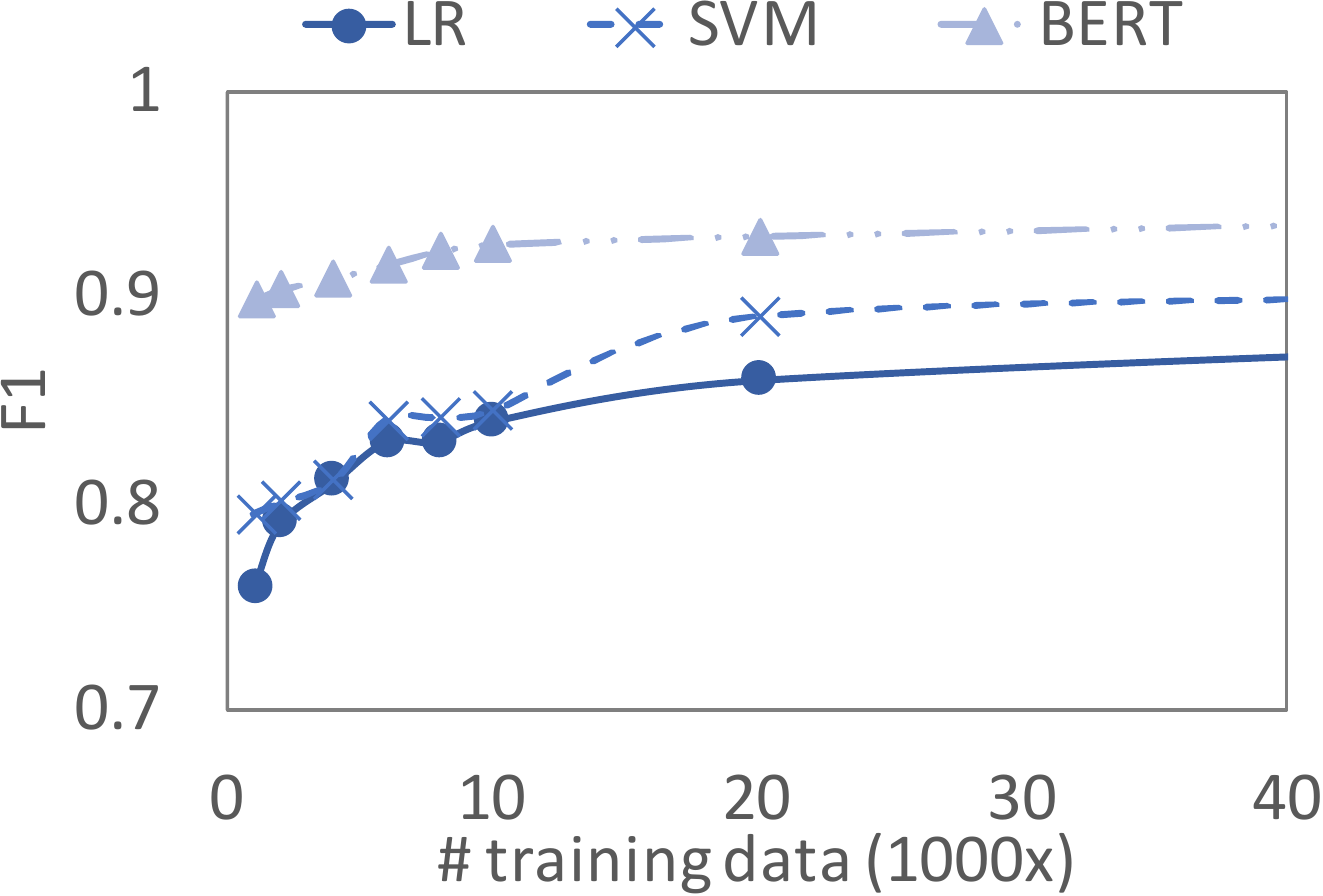}
	    \caption{AMAZON}
	    \label{fig:datasize_f1_amazon}				
	\end{subfigure}	
    \begin{subfigure}{0.24\linewidth}
	    \includegraphics[width=\linewidth]{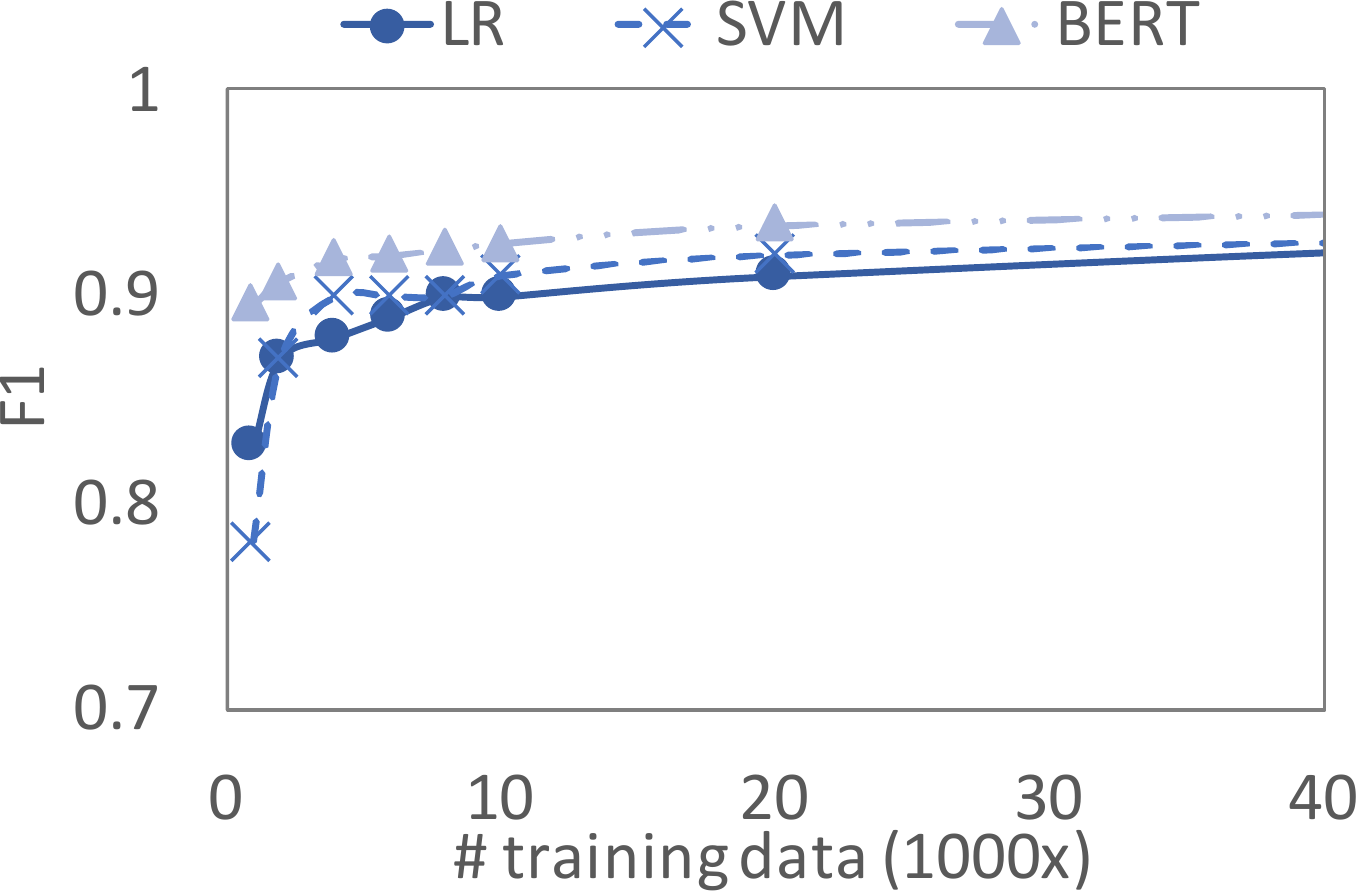}
	    \caption{YELP}
	    \label{fig:datasize_f1_yelp}				
	\end{subfigure}	
    \begin{subfigure}{0.24\linewidth}
	    \includegraphics[width=\linewidth]{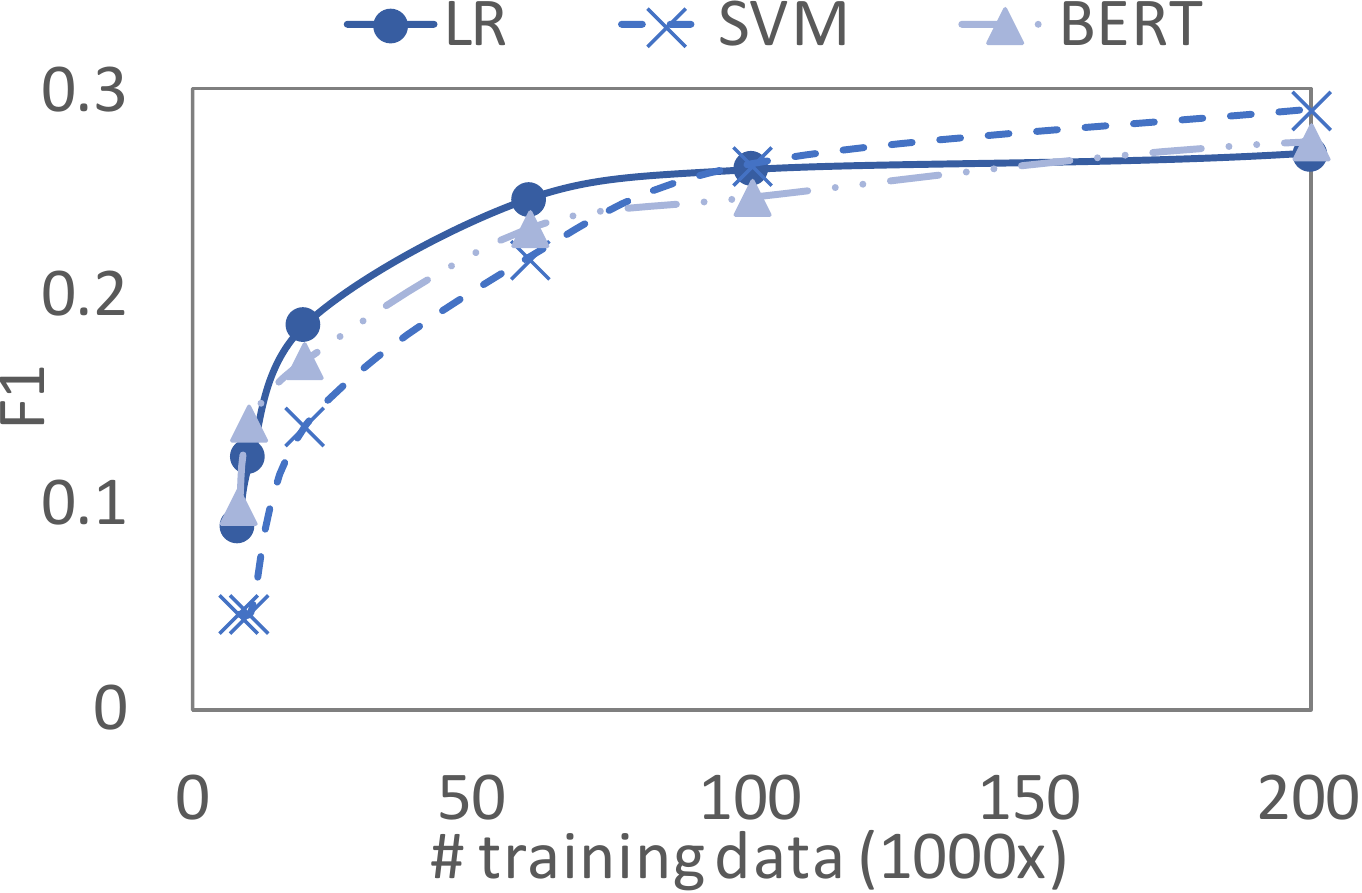}
	    \caption{FUNNY}
	    \label{fig:datasize_f1:funny}				
	\end{subfigure}	
    \begin{subfigure}{0.24\linewidth}
	    \includegraphics[width=\linewidth]{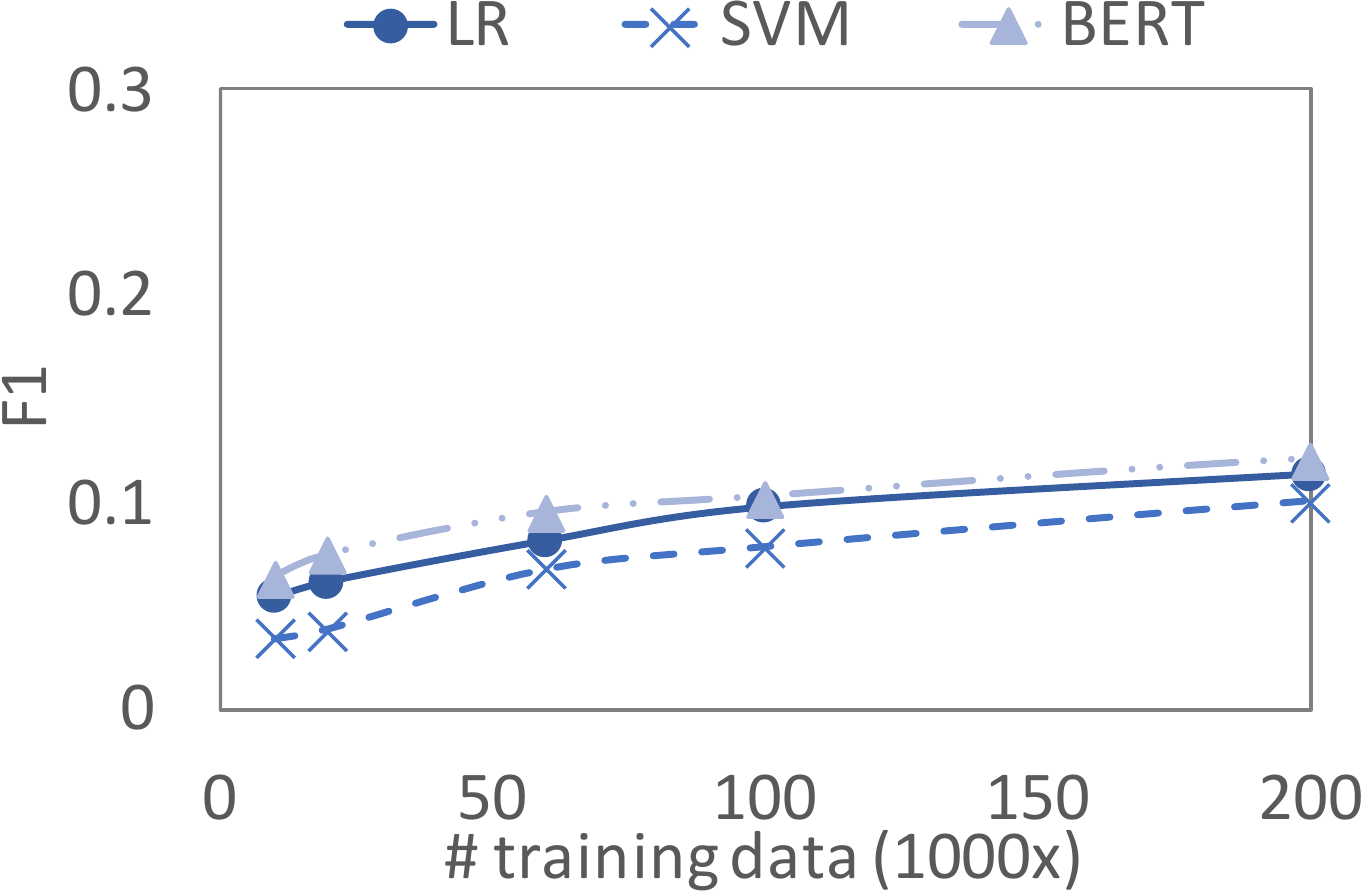}
	    \caption{BOOK}
	    \label{fig:datasize_f1:book}				
	\end{subfigure}		
	\caption{F1 of LR, SVM, and BERT by increasing the size of training data}
	\label{fig:datasize_f1}	
	\vspace{-2mm}
\end{figure*}		

%\subsubsection{Effects of training size}
\subsubsection{Size of training set} \label{subsec:datasize}%larger size, f1
% setting
% \how much F1 improvements. Can we rely on enlarging training set to improve F1?
%
To understand how the size of training set affects tagging quality, we increase the number of training data for LR, SVM, and BERT and measure their F1s. We use AMAZON, YELP, FUNNY, and BOOK as representative datasets, given that they contain abundant labels. We fix the number of test data to 100,000, since we do not observe significant differences in tagging quality when using more test data. 

% F1 flattens, gap narrows, increasing the training size benefits LR/SVM more than BERT
We plot the F1s of LR, SVM, and BERT with regard to the increasing numbers of training data in Figure~\ref{fig:datasize_f1}. As expected, all models obtain F1 improvements when getting more training data, especially for LR and SVM. %By increasing the training size of AMAZON from 2000 to 20000, the F1 of LR, SVM, and BERT are improved from 0.79, 0.80, and 0.90 to 0.86, 0.89, and 0.93, respectively. 
Increasing the training size of AMAZON from 2000 to 20,000 improves the F1 of LR (from 0.79 to 0.86), SVM (from 0.80 to 0.89), and BERT (from 0.90 to 0.93), respectively. The F1 improvements of LR and SVM (0.07 and 0.09) are greater than BERT (0.03). Similarly, the F1 improvements of LR and SVM on YELP (0.04 and 0.05) are higher than BERT (0.02), suggesting that increasing training size to promote F1 is more effective on simple models than deep models. In other words, the performance gap between simple models and deep models narrows with increase in training size.
%In other words, both LR and SVM are catching up with BERT given more training data, which explains their minor F1 differences on AMAZON and YELP in Figure~\ref{fig:f1_high_p}. 
On FUNNY and BOOK, all models achieve smaller tagging F1s than that they achieve on AMAZON and YELP. To better show the differences, we set the maximal F1s to 0.3 as shown in Figure~\ref{fig:datasize_f1:funny} and Figure~\ref{fig:datasize_f1:book}. No matter whether the training size is small or large, BERT does not show a superior F1 compared with LR and SVM. Along with the increase of training size, BERT performs similarly to LR and even worse than SVM on FUNNY, whereas it performs similarly to LR and slightly better than SVM on BOOK. %\yuliang{The content looks fine, but the language still needs some work. } \jinfeng{I have fixed the typos}

% increasing number of training data to increase F1. But this is less effective for large dataset and for dirty dataset. (improvement less than 0.03)

% Our analysis in this section suggests that one, 
Notably, although LR and SVM achieve similar or even better F1s than BERT in some cases, they require numerous labels at the scale of tens of thousands. This is a big concern since data labeling may rely on human annotation rather than rule generation. Given that most of real-world datasets are on a small scale, BERT is still an appealing option.  \\

% \begin{table}[!ht]
% \centering
% \begin{tabular}{|l|l|l|l|}
% \hline
% Dataset & overall   & positive & negative \\ \hline
% SUGG    & 10,466    & 5,000    & 9,042    \\ \hline
% HOTEL   & 7,535     & 1,604    & 7,265    \\ \hline
% SENT    & 8,010     & 2,093    & 7,774    \\ \hline
% PARA    & 8,212     & 2,899    & 7,714    \\ \hline
% FUNNY   & 571,841   & 147,344  & 538,163  \\ \hline
% HOMO    & 5,059     & 4,180    & 1,730    \\ \hline
% HETER   & 5,034     & 4,299    & 1,462    \\ \hline
% TV      & 20,209    & 13,745   & 14,448   \\ \hline
% BOOK    & 373,198   & 84,625   & 363,056  \\ \hline
% EVAL    & 8,357     & 4,896    & 7,114    \\ \hline
% REQ     & 8,357     & 3,659    & 7,656    \\ \hline
% FACT    & 8,357     & 5,699    & 6,533    \\ \hline
% REF     & 8,357     & 551      & 8,225    \\ \hline
% QUOTE   & 8,357     & 1,073    & 8,260    \\ \hline
% ARGUE   & 21,648    & 14,309   & 17,078   \\ \hline
% SUPPORT & 21,648    & 9,554    & 19,979   \\ \hline
% AGAINST & 21,648    & 10,975   & 19,255   \\ \hline
% AMAZON  & 1,001,083 & 674,821  & 605,106  \\ \hline
% YELP    & 232,282   & 162,055  & 158,158  \\ \hline
% \end{tabular}
% \caption{Dataset vocabulary}
% \label{table:vocabulary}
% \end{table}

\begin{figure}[t]
    \centering
    \scalebox{0.7}{
    \includegraphics[width=\linewidth]{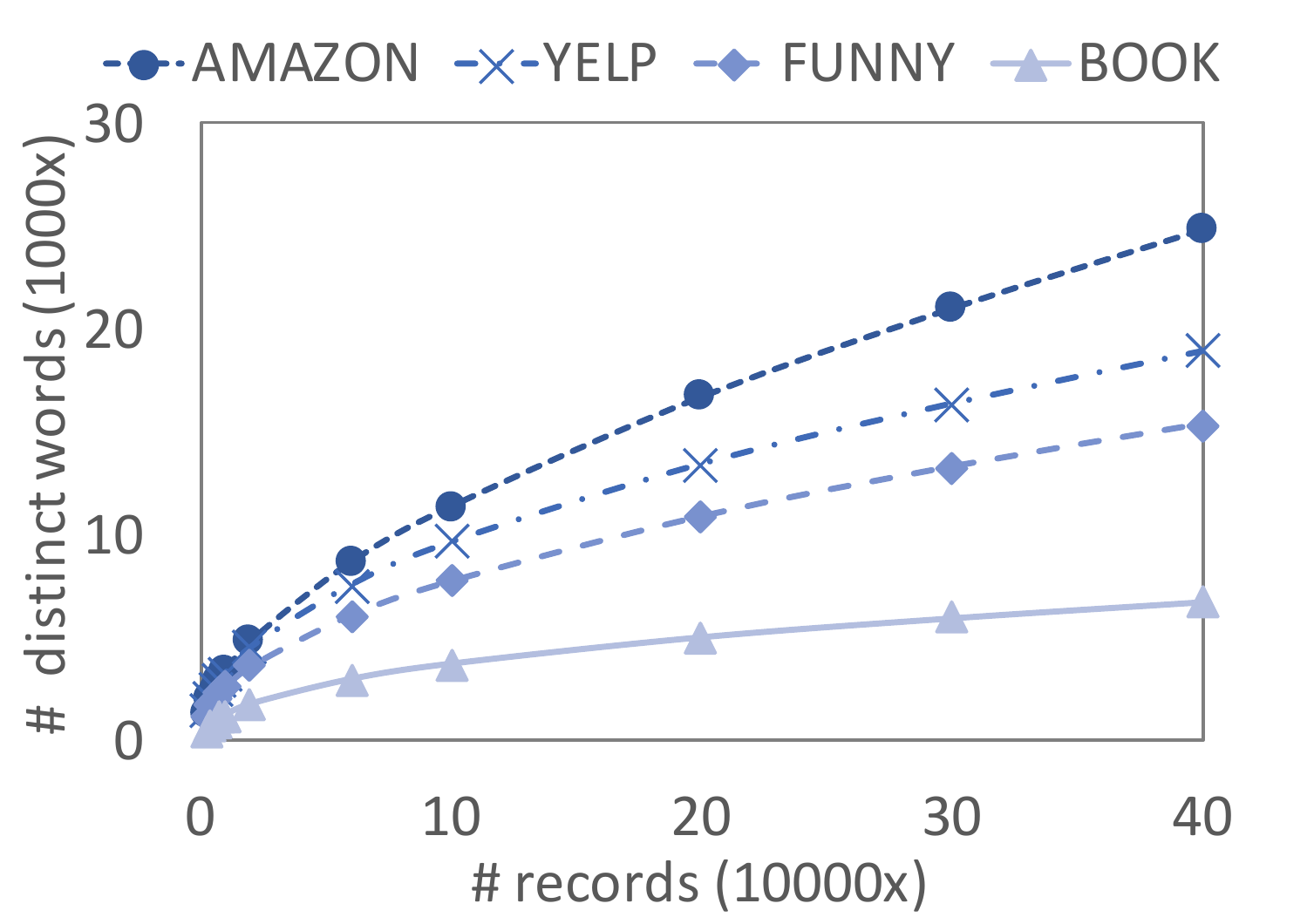}
    }
    \caption{The number of distinct words keeps increasing when expanding the training set}
    \label{fig:vocab_scale}
    \vspace{-3mm}
\end{figure}

\noindent\textbf{Vocabulary size analysis.} To explain that why enlarging the training set increases the tagging quality, we calculate the numbers of distinct words (i.e. vocabulary size) as we feed more records to expand a training set. We depict the increment of distinct words in Figure~\ref{fig:vocab_scale}. On each of AMAZON, YELP, FUNNY, and BOOK, the number of distinct words keeps increasing, leading to more words being included in training processes.  This explains why F1 becomes higher when the training size increases.  
We also observed that F1s of all three models change negligibly after the training size reaches a certain number (e.g., 100,000 on YELP), indicating that the promoting effects of training size on F1 are limited when the sizes reach certain thresholds.

\begin{figure*}[!ht]
    \centering
    \begin{subfigure}{0.24\linewidth}
	    \includegraphics[width=\linewidth]{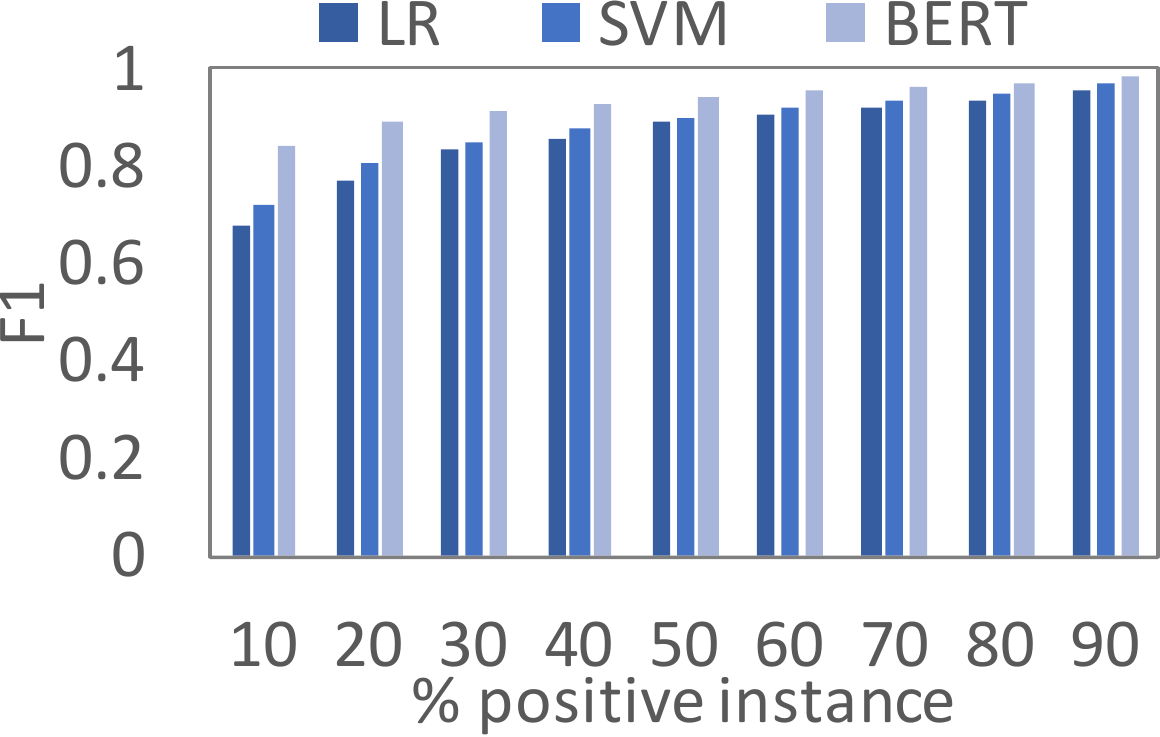}
	    \caption{AMAZON}
	    \label{fig:ratio_f1_amazon}				
	\end{subfigure}	   
    \begin{subfigure}{0.24\linewidth}
	    \includegraphics[width=\linewidth]{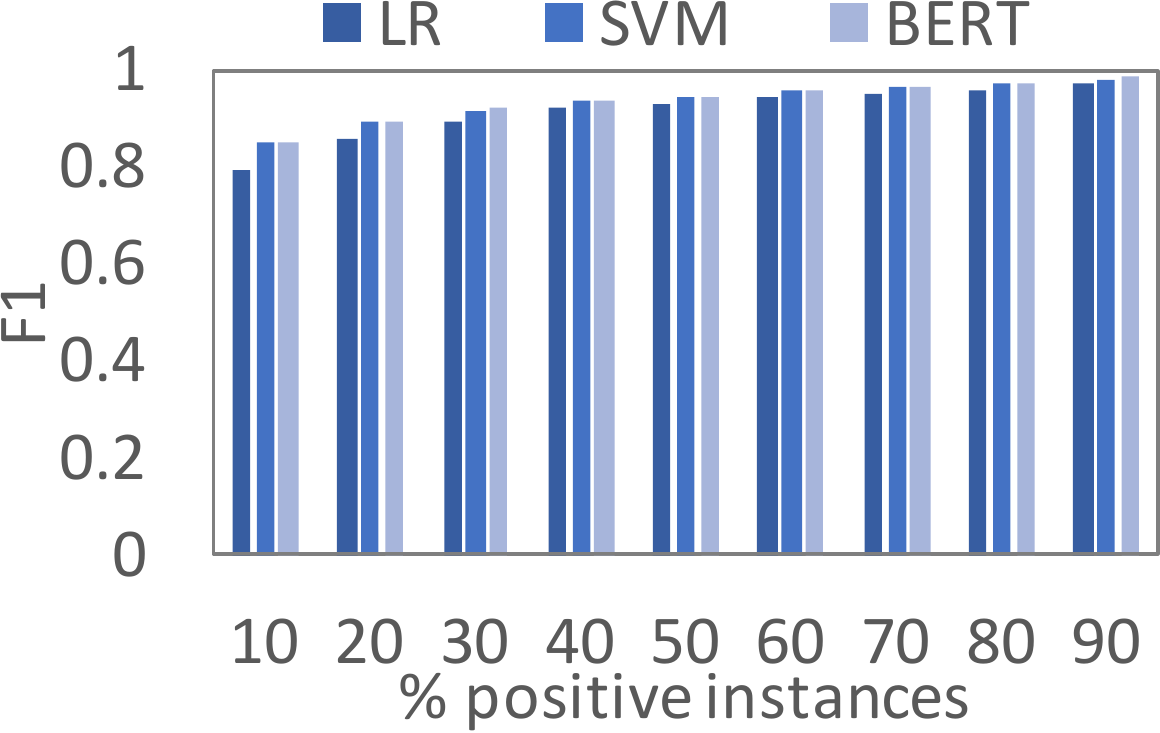}
	    \caption{YELP}
	    \label{fig:ratio_f1_yelp}				
	\end{subfigure}	   	
    \begin{subfigure}{0.24\linewidth}
	    \includegraphics[width=\linewidth]{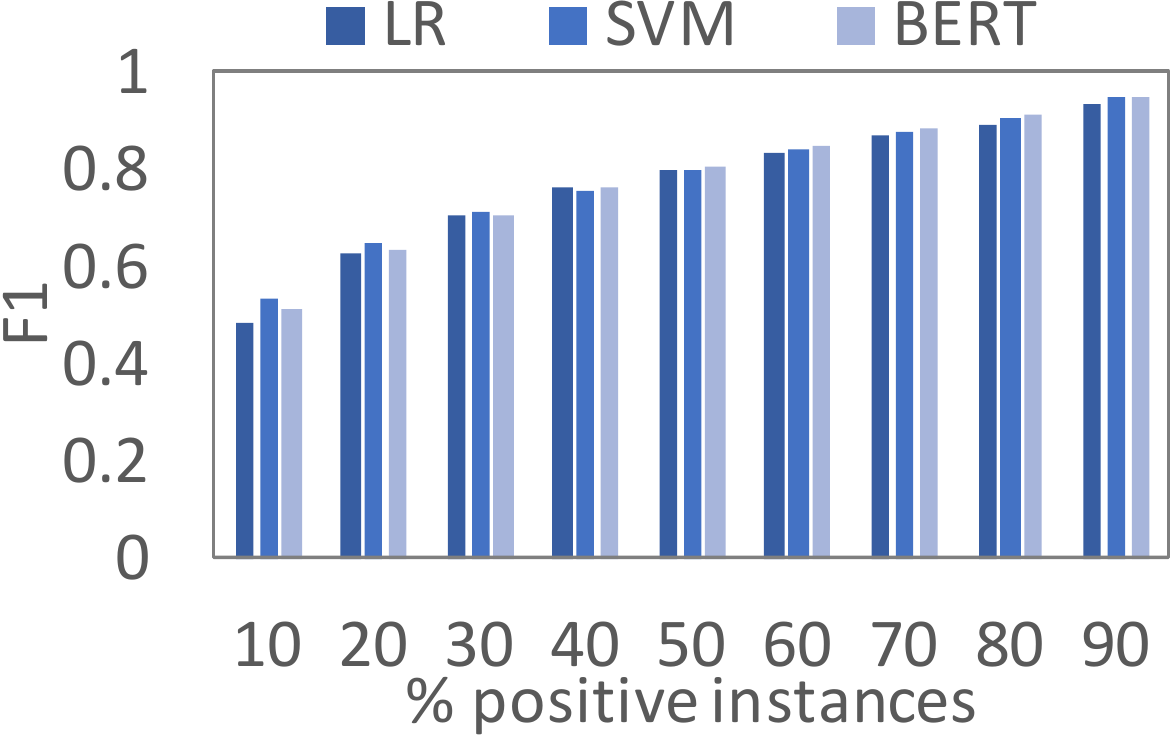}
	    \caption{FUNNY}
	    \label{fig:ratio_f1_funny}				
	\end{subfigure}	    
    \begin{subfigure}{0.24\linewidth}
	    \includegraphics[width=\linewidth]{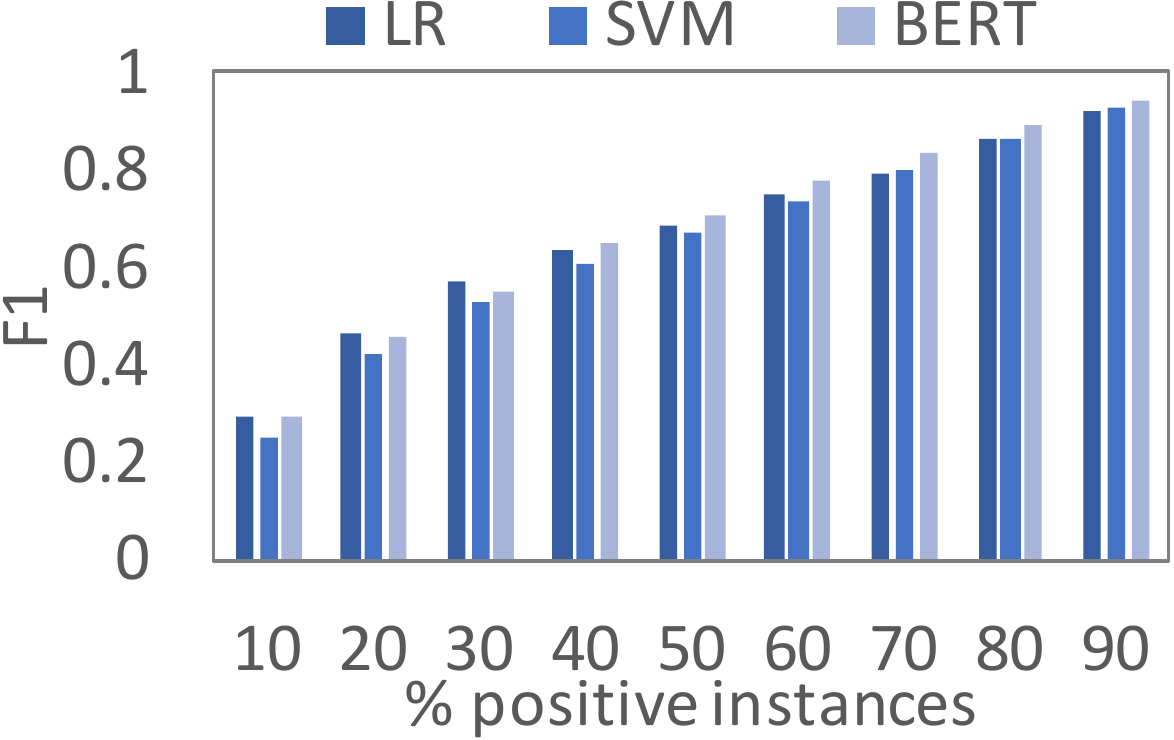}
	    \caption{BOOK}
	    \label{fig:ratio_f1_book}				
	\end{subfigure}	
	\caption{F1 of LR, SVM, and BERT on the 4 large datasets sub-sampled (100k) with different label ratios.}
	\label{figure:ratio_f1}
\end{figure*}	

\begin{table*}[!t]
\caption{Informative tokens regarding each dataset and their frequencies of occurrence in \textbf{P}ositive and \textbf{N}egative instances. We sorted all tokens by $\textbf{P}-\textbf{N}$ descendingly and presented the top 5 tokens. Top tokens on AMAZON and YELP are all sentiment words. Top tokens on FUNNY* and BOOK* include stop words such as ``that''.}
\label{table:marker_tokens}
\centering
\scalebox{1}{
\begin{tabular}{ccc|ccc|ccc|ccc}
\toprule
AMAZON & \textbf{P}    & \textbf{N}     & YELP      & \textbf{P}    & \textbf{N}    & FUNNY* & \textbf{P}    & \textbf{N}     & BOOK* & \textbf{P}    & \textbf{N}    \\ 
\midrule
great  & 0.27 & 0.09 & great     & 0.39 & 0.15 & that  & 0.75 & 0.41 & he   & 0.13 & 0.06 \\
love   & 0.15 & 0.05 & delicious & 0.14 & 0.02 & you   & 0.69 & 0.37 & that & 0.25 & 0.19 \\
best   & 0.12 & 0.04 & love      & 0.17 & 0.06 & on    & 0.70 & 0.40 & she  & 0.13 & 0.08 \\
easy   & 0.08 & 0.02 & friendly  & 0.16 & 0.06 & of    & 0.87 & 0.58 & her  & 0.13 & 0.08 \\
well   & 0.16 & 0.11 & best      & 0.19 & 0.09 & at    & 0.60 & 0.31 & him  & 0.06 & 0.03 \\
\bottomrule
\end{tabular}
}
\end{table*}

\subsubsection{Label ratio}
% \subsubsection{Effects of label ratio} \label{subsec:ratio} % more P, f1
We evaluate how the ratio of positive instances affects the F1s of tagging models. To perform the evaluation, we adopt AMAZON, YELP, FUNNY, and BOOK as training datasets and adjust their ratios of positive instances from 10\% to 90\%. For each ratio $r$, we randomly sample 100,000 records from each dataset. Within these 100,000 records, the number of positive instances is $r$ of 100,000 and the number of negative instances is $1-r$ of 100,000. The records are randomly split into two parts, of which 80\% of the records are devoted to training set and 20\% are devoted to testing set. 
%\yuliang{it's unclear how the train/test sets are generated.}\jinfeng{I rewrote}
As a result, all models obtain higher F1s on 4 datasets when the ratios of positive instances increase, suggesting that higher ratios of positive instances bring tagging models towards higher F1s. This result is shown in Figure~\ref{figure:ratio_f1}. 

We find that F1 improvements regarding the increase in the ratio of positive instances exhibit two trends. First, they are more significant when the ratio of positive instances is small ($< 25\%$), while more flattened when the ratio is large ($\geqslant 25\%$). For example, F1 of BERT gains 0.06 improvement on AMAZON when the ratio increases from 10\% to 20\%, but only 0.01 when the ratio changes from 80\% to 90\%. Second, F1 improvements on AMAZON and YELP are less obvious than those on FUNNY and BOOK. For example, when the ratio increases from 10\% to 20\%, F1 improvements of BERT on AMAZON and YELP are 0.05 and 0.04, while F1 improvements on FUNNY and BOOK are 0.12 and 0.16. This result suggests that increasing label ratio to promote F1 is more effective on dirty datasets (FUNNY and BOOK) than clean datasets (AMAZON and YELP). 

%We also find when the ratio of positive labels is higher, the F1 is higher too. This applies to all tagging models. In fact, when the ratio is higher, the F1 for random guessing will be higher too. 

Besides F1 improvements, we also observe that the F1 gaps between LR/SVM and BERT decrease when the ratio of positive instances increases. On AMAZON, F1 gap between LR and BERT decreases from 0.16 to 0.03 when the ratio increases from 0.1 to 0.9. Similarly, F1 gap on YELP decreases from 0.06 to 0.01 when the ratio goes up. The corresponding decreases in F1 gap may due to more unseen positive instances that are brought by the dataset with an increasing ratio. These unseen positive words have more influences on LR and SVM than on BERT as LR and SVM rely more on the occurrence of words for the tagging task.

% \subsubsection{Analysis on informative tokens} 
\subsubsection{Informative tokens} 
The F1s of a tagging model on different datasets can be significantly various, even when these datasets have the same training size and label ratio. For example, both AMAZON and BOOK* have more than 1 million labeled instances and present a percentage of positive instances of 50\%.  However, BERT achieves 0.96 F1 on AMAZON and 0.74 F1 on BOOK* (shown in Figure~\ref{fig:f1_high_p}), between which the F1 gap is 0.22. We also observe similar behaviors of other models on different datasets, such as the F1 gap of SVM on YELP and FUNNY* is 0.15.

We assume that tagging labels on FUNNY* and BOOK* are harder because these two datasets are dirtier in labels. To understand what ingredients in a dataset make it hard to be tagged, we analyze informative tokens that can separate positive records from negative records. First, we try to identify such tokens from the datasets. To perform the identification, we calculate the percentage of positive instances containing token $t$ (denoted as $P$) and the percentage of negative instances containing $t$ (denoted as $N$). $P$ and $N$ measure the occurrence of $t$ in records concerning two different types of labels. We then sort the $P - N$ values of all tokens in AMAZON, YELP, FUNNY*, and BOOK* in descending order. Thereafter, we present the top 5 informative tokens and their frequencies of occurrence in table~\ref{table:marker_tokens}. The result shows that top tokens from AMAZON and YELP are all sentiment words that express positive opinions, such as ``great'' and ``love''. These words semantically link to the task itself, i.e. tagging positive sentences. However, top tokens from FUNNY* and BOOK* contain some stop words such as ``that'' and ``on''. These stop words appear in high frequencies in both positive and negative records.

\begin{figure}[t]
    \centering
    \scalebox{0.95}{
    \includegraphics[width=\linewidth]{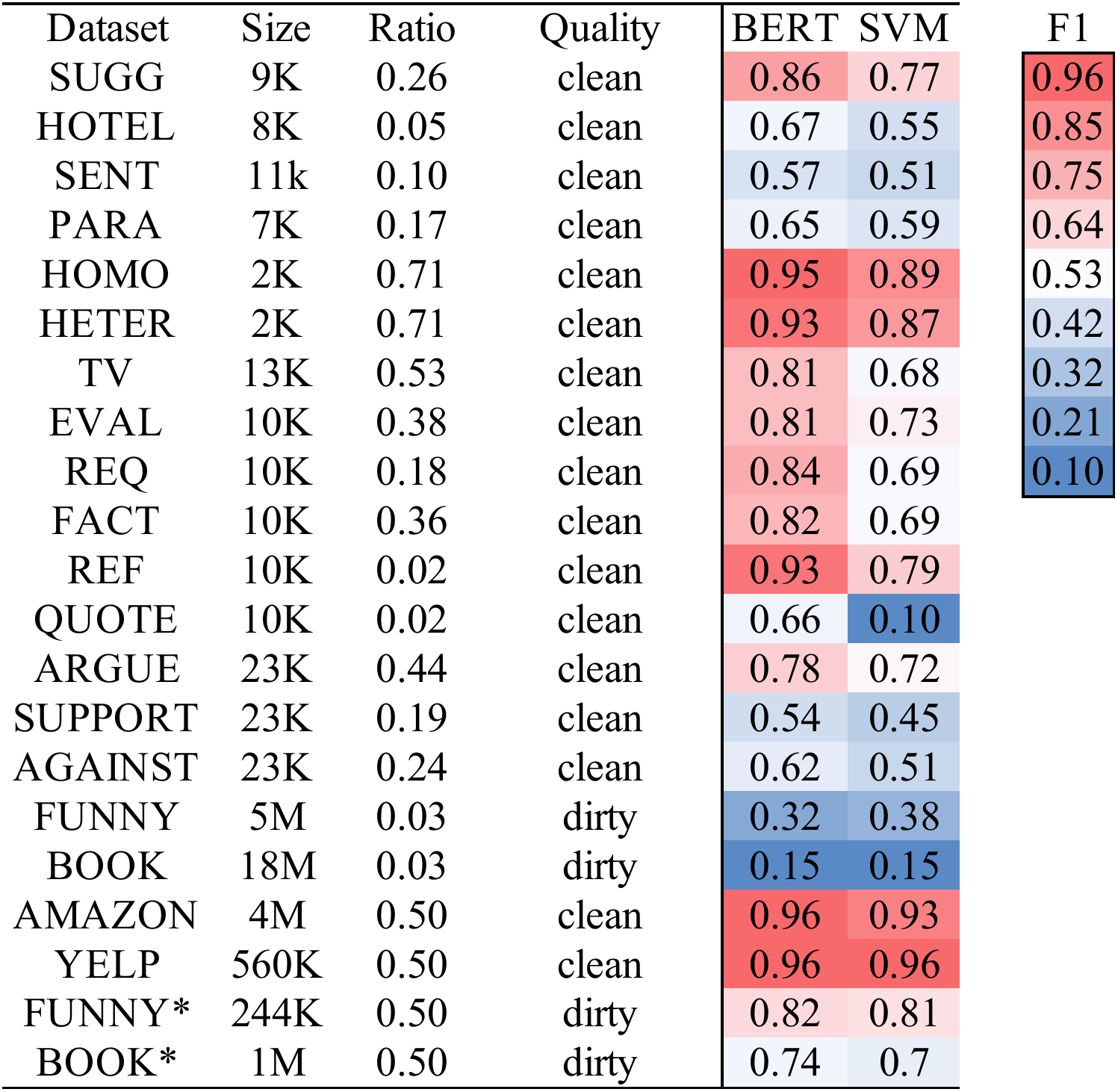}
    }
    \caption{F1s of BERT and SVM on 21 different datasets}
    \label{fig:heatmap}
\end{figure}

\subsection{Towards a higher F1} \label{subsec:heatmap}% how to use our results

% heat map (why? to effectively improve tagging F1s
Our study can serve as a reference for practitioners to understand how F1 is affected. To make sure that our study can be utilized, we summarize and visualize our results as a heat map (Figure~\ref{fig:heatmap}). We select BERT and SVM as the representative deep model and simple model, respectively. We describe their training size, ratio of positive instances, label quality of individual dataset, and tagging F1s in Figure~\ref{fig:heatmap}. To visualize F1 values, we present small F1 ($< 0.53$) as blue color and deepen color when values decrease, and present large F1 ($\geqslant 0.53$) as red color and deepen color when values increase. By retrieving our heat map, practitioners can estimate the approximate tagging F1s for their datasets and choose the appropriate tagging model to obtain better tagging F1s. 
%If a dataset looks like AMAZON and YELP, the practitioner can expect a similarly high F1 regardless of tagging models.

% BERT improvement
Our study indicates that practitioners should try BERT if their targeting datasets are small. They can expect F1 improvement as much as 0.56. However, F1 improvement of BERT will be less significant if a dataset has a large number of labels. When tagging the datasets with both large sizes and high ratios of positive instances, such as FUNNY, BOOK, AMAZON, YELP, FUNNY*, and BOOK*, BERT does not show appealing F1 improvements while taking plenty of training time (several days). In this case, practitioners may consider simple models like SVM as an alternative choice, given that SVM can achieve a similar F1 value while taking much less training time, in comparison to BERT. 

% Dataset preparation
In addition to choosing appropriate tagging models, practitioners should also pay attention to dataset selection. Our study shows that AMAZON and YELP enable high F1s regardless of tagging models. These two datasets own abundant training data, exhibit balanced ratios, and include clean labels. HOMO and HETER also allow tagging models to achieve high F1s. Although these two datasets are smaller in size compared with AMAZON and YELP, they have more positive instances ($> 70\%$) than negative instances. %The positive skewness lowers tagging difficulties.
The higher label ratio makes the semantic tagging easier. In contrast, some datasets like FUNNY and BOOK can hinder tagging models from obtaining high F1s. The reason is that their labels are dirty and ratios of positive instances are small. Therefore, practitioners should be cautious when preparing datasets and try to get datasets with large size, high ratio of positive instances, and pure cleanliness (e.g. no missing annotations).

%A practitioner can expect BERT to achieve a higher F1 than simple models on small datasets (e.g. QUOTE and HOTEL). while BERT offers limited F1 improvements on large datasets (e.g. AMAZON and YELP) yet consumes significant training time. If the practitioner still feel unsatisfied with the F1 that BERT can provide, he can look into dataset characteristics. He should expect a low F1 if his dataset has sufficient labels (like SENT and PARA), skews against positive instances (like FUNNY and BOOK) and lacks informative tokens (like FUNNY* and BOOK*). He may optimize his dataset in one or more aspects towards sufficient labels like AMAZON and YELP, a higher percentage of positive instances like HOMO and HETER or better informative tokens (like REF and REQ). 

%% file: conclusion.tex
\vspace{-2mm}
\section{Conclusion} \label{sec:conclusion}

Our study is the most comprehensive one that used the largest number of real-world datasets to compare deep models and simple models. Our results reveal for the first time that dataset characteristics are the key factors to determine whether deep models can achieve better tagging quality than simple models. Given the raw complexity of real-world datasets, choosing a suitable tagging model for a specific dataset rather than sticking with deep models should be the way of performing tagging tasks in the future. Our study, especially the visualized heat map will be the most informative instruction for practitioners to choose a suitable tagging model for their dataset, by considering its scale, label ratio, and cleanliness. 

%% file: acknowledgement.tex
\section{Acknowledgments}
We thank Alon Halevy for his discussions in the early stages of this work. We also thank Nikita Bhutani for giving numerous comments that help significantly improve the paper.

%% file: appendix.tex
\section*{APPENDIX}

%\xiaolan{Maybe consider adding section numbers for easier reference? I noticed edits were highlighted in the main paper, but not in the Appendix. Maybe also highlights edits in the Appendix?}

\section*{Effect of Calibration on Imbalanced Datasets}
% Calibration improve both BERT's and simple model's F1s. 
In the experiments, We use argmax that takes the larger output score as the final label of a data point, recalling that each data point has two output scores with regard to the positive and the negative accordingly. Our evaluations follow the same post-processing of outputs for different models. 

Compared with argmax, calibration can effectively tackle label imbalance. To determine whether calibration affects the comparison of simple models and deep models on large imbalanced datasets, we calculate the maximum F1 of a model by varying the calibration threshold on the positive scores. Specifically, We fix the number of thresholds and sample thresholds from the range of maximum and minimum scores. We calculate F1 regarding each threshold and take the maximum F1 as the final F1. 

We compare BERT and LR/SVM on FUNNY and BOOK, the two largest imbalanced datasets. We randomly sample 1 million labels for experiments such that BERT can finish training within 24 hours. We fix the number of thresholds to 100, 200, 300, and 400, respectively. The results are presented in Figure~\ref{fig:calibrate}. After we tackled imbalance, simple models still show better F1s on FUNNY and similar F1s on BOOK, in comparison to BERT. This is consistent with our previous conclusion. We did see that adding calibration can improve F1s of all models, especially BERT, but does not change the tendency that simple models perform similar or even better than BERT.

\begin{figure}[!ht]
    \centering
    \begin{subfigure}{0.49\linewidth}
	    \includegraphics[width=\linewidth]{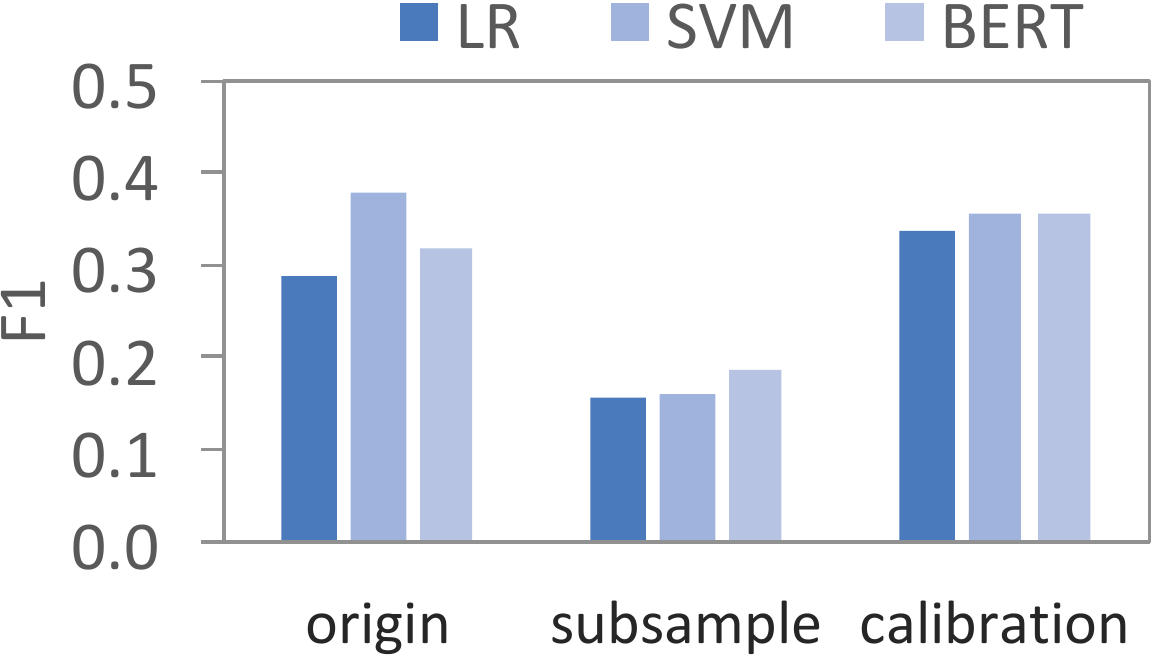}
	    \caption{FUNNY}
	    \label{fig:subsample_funny}				
	\end{subfigure}	
    \begin{subfigure}{0.49\linewidth}
	    \includegraphics[width=\linewidth]{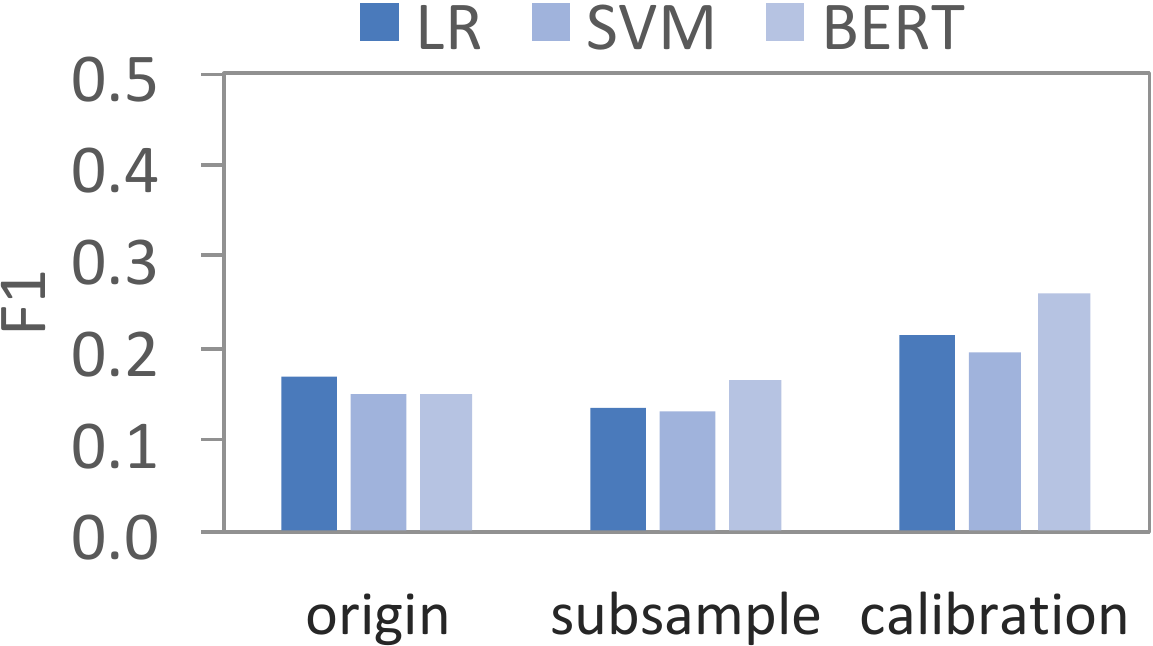}
	    \caption{BOOK}
	    \label{fig:subsample_book}				
	\end{subfigure}	    
    \caption{F1 of LR, SVM, BERT on subsampled FUNNY/BOOK}
    \label{fig:subsample}
\end{figure}

\section*{Effect of Subsampling on Imbalanced Datasets}
Since LR/SVM can achieve better F1s than BERT on FUNNY and BOOK as shown in Figure~\ref{fig:f1_low_p}, we investigate more about their comparison on imbalanced datasets by applying subsampling.  

We conduct subsampling experiments that adjust the ratio of train set only. We keep the ratio of test set unchanged. On train set, we undersample negative labels of FUNNY and BOOK to keep the label ratios to 50\%. After undersampling, FUNNY/BOOK has 195351/911169 training samples, respectively. We present F1’s as well as calibrated F1’s for LR, SVM, and BERT. We also report F1’s on original datasets for comparison. As shown in Figure~\ref{fig:subsample_funny}, after using subsampling, F1s would change correspondingly. If we do subsampling without calibration, simple models show slightly worse F1s than BERT. However, if we add calibration to subsampling, then the F1s achieved by simple models become the same as BERT on FUNNY. Overall, different methods indeed affect the output of models but do not change the main statement of our study.

%As shown in Figure~\ref{fig:subsample_funny}, LR/SVM/BERT achieve worse F1s on subsampled dataset than those on original FUNNY dataset. This is because all models learn argmax on train set yet apply it to test set that has a very different label ratio. After we calibrate outputs to find maximum F1s by varying thresholds on the test set, all models achieve better F1s. However, no model appears as a clear winner on FUNNY. We present the results of BOOK in Figure~\ref{fig:subsample_book}. SVM/LR outperform BERT on subsampled dataset but after calibration, BERT appears as a winner. The F1 gain is 0.05, which is smaller than any F1 gains offered by BERT on smaller datasets (see Figure 1 and 2). Overall, there is not a model that consistently outperforms the others under all circumstances. 

We present subsampling experiments that adjust the label ratios of train set and test set simultaneously, as discussed in Section 6.2.2. Our experiments reference the setting of Imbalanced-learn~\cite{imbalanced_learn} that subsamples both train set and test set to keep the same label ratio. Imbalanced-learn is a famous Sklearn library that receives 4.4k stars on Github. In the experiments, we oversample FUNNY and BOOK and adjust their ratios of positive instances from 10\% to 90\%. We report the F1s in Figure~\ref{fig:ratio_f1_funny} and Figure~\ref{fig:ratio_f1_book}. The results show that both LR/SVM and BERT obtain higher F1s when the ratio increases. However, LR or SVM only outperforms BERT in some cases, which means there is no clear winner regarding F1. However, if we further take runtime into account, LR and SVM have the advantage of faster runtime. Therefore, even we consider subsampling when comparing F1s, our main statement remains unchanged that simple models achieve similar tagging quality to deep models on large datasets, but the runtime of simple models is much shorter.

\begin{figure}[!ht]
    \centering
    \scalebox{0.7}{
	    \includegraphics[width=\linewidth]{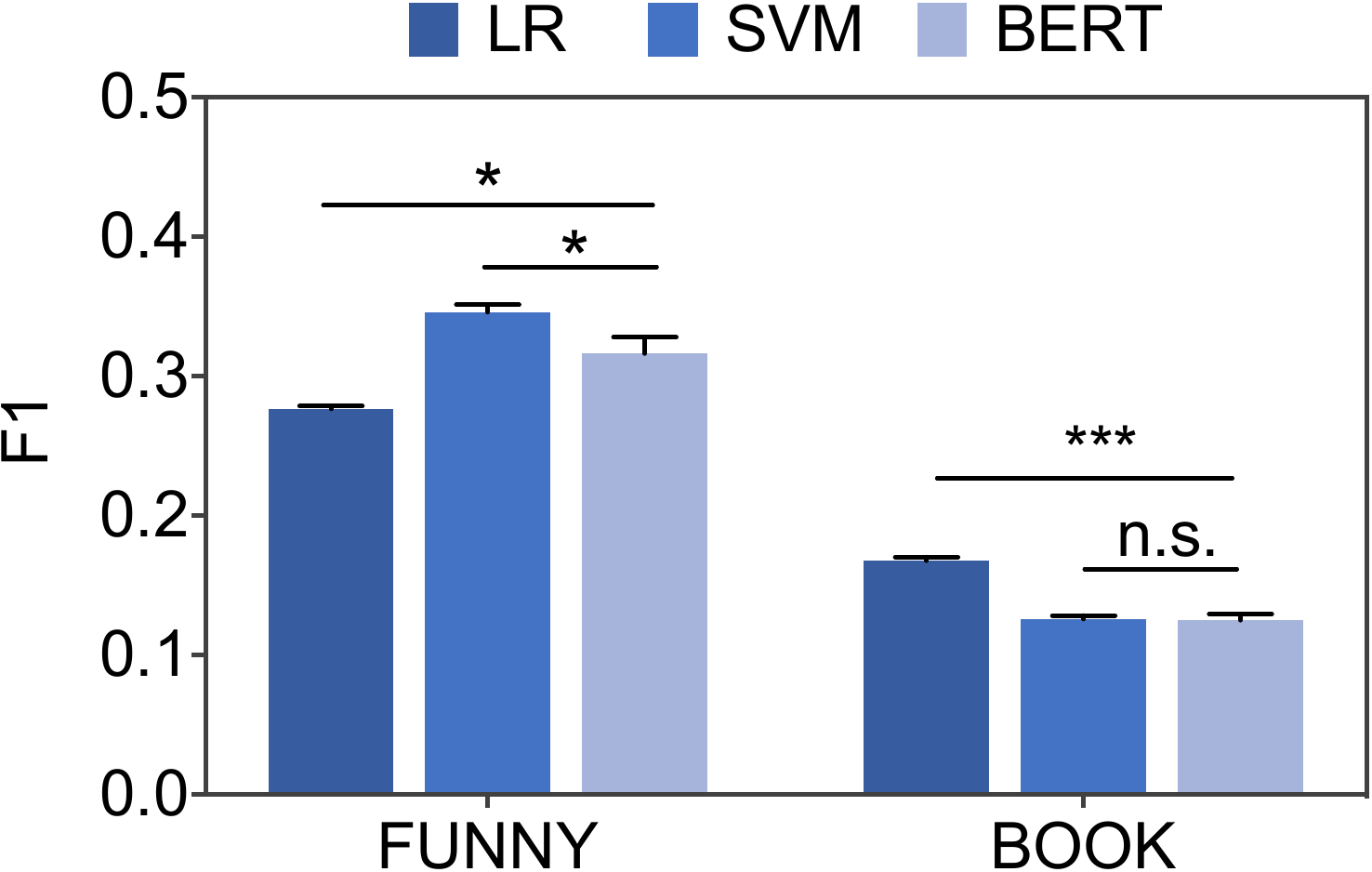}
	 }
    \caption{Error bar and statistical significance of LR, SVM, and BERT on FUNNY/BOOK (1 million labels)}
    \label{fig:error_bar}
\end{figure}

\section*{Effect of Randomness}
We have conducted more experiments to evaluate the randomness and statistical significance. We set different random seeds to shuffle the datasets and repeat the process for 3 times and run LR, SVM, and BERT on Funny and BOOK with 1 million labels each. We perform comparison with LR vs BERT or SVM vs BERT and perform the statistical analysis using GraphPad Prism 7. The data are presented as mean  SD, n = 3. $P$ value is generated by the Student’s t test (n.s., not significant at $P > 0.05$, $*P < 0.05$, $**P < 0.01$, $***P <0.001$).
 
As shown in Figure~\ref{fig:error_bar}, when evaluating F1s of LR, SVM, and BERT on 1 million labels from FUNNY, SVM performs better than BERT ($P < 0.05$) while LR shows worse performance than BERT ($P < 0.05$), suggesting that at least one simple model outperforms BERT. Further, on 1 million labels from BOOK, we observe a much stronger performance of LR ($P < 0.001$) while SVM keeps a comparable performance compared with BERT, strengthening the presumption that simple model can achieve the same, in some conditions even better, performance compared with deep model.

\section*{Effect of Pre-training Embeddings on Simple Models}

\begin{figure*}[t]
    \centering
	    \includegraphics[width=\linewidth]{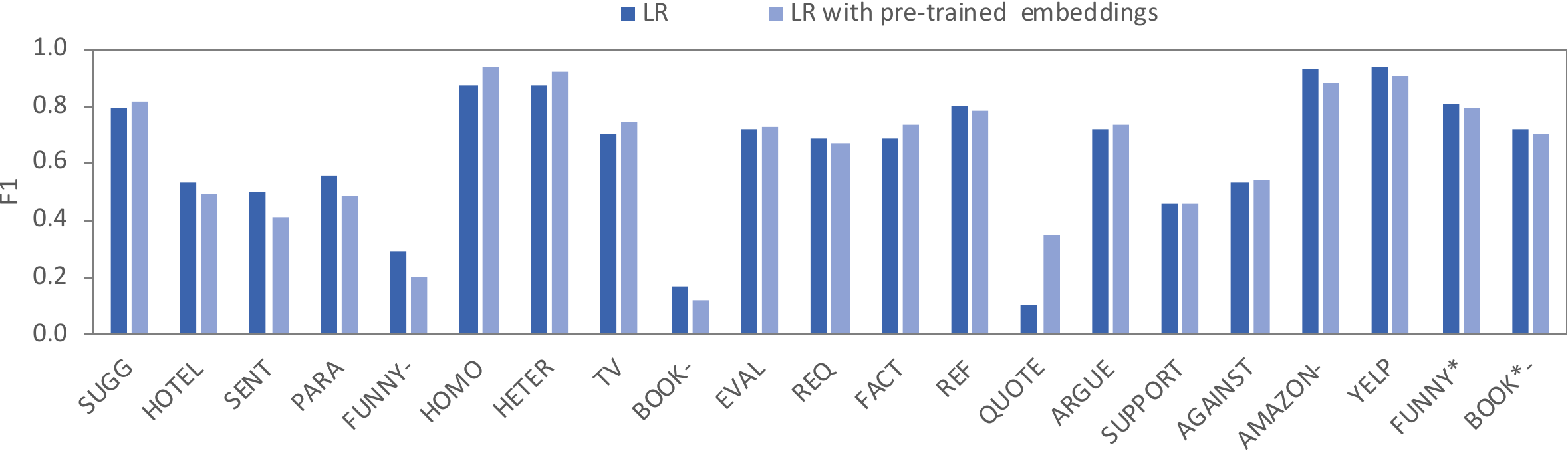}
	    \caption{F1 of LR when using BERT pre-training embeddings as featurization vectors}
	    \label{fig:pre_train_lr}	
	    \vspace{4mm}
	    \includegraphics[width=\linewidth]{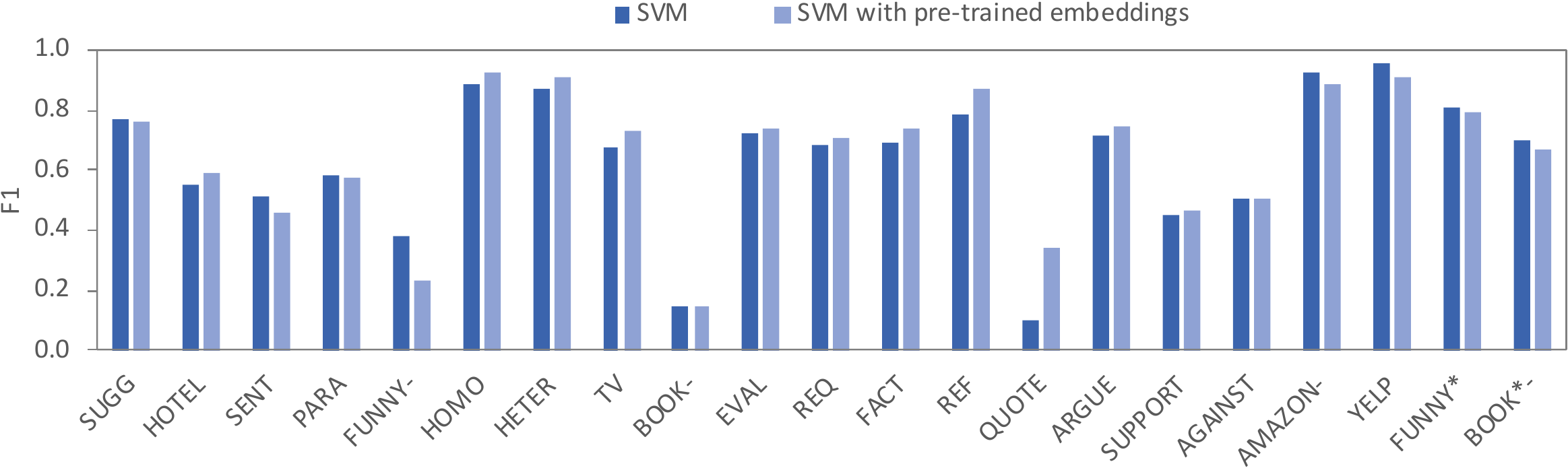}
        \caption{F1 of SVM when using BERT pre-training embeddings as featurization vectors}
	    \label{fig:pre_train_svm}	
	    	    \vspace{4mm}
	    \includegraphics[width=\linewidth]{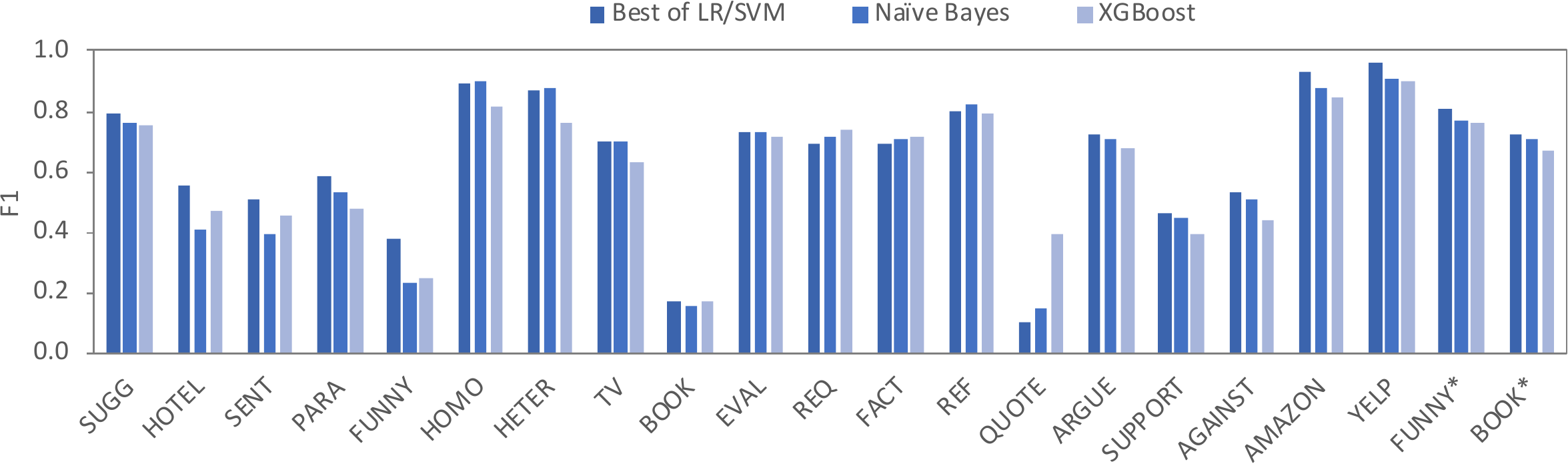}
        \caption{F1 of the best of LR/SVM, Naive Bayes, and XGBoost}
	    \label{fig:naive_bayes_xgb}	
	    	    \vspace{4mm}
	    \includegraphics[width=\linewidth]{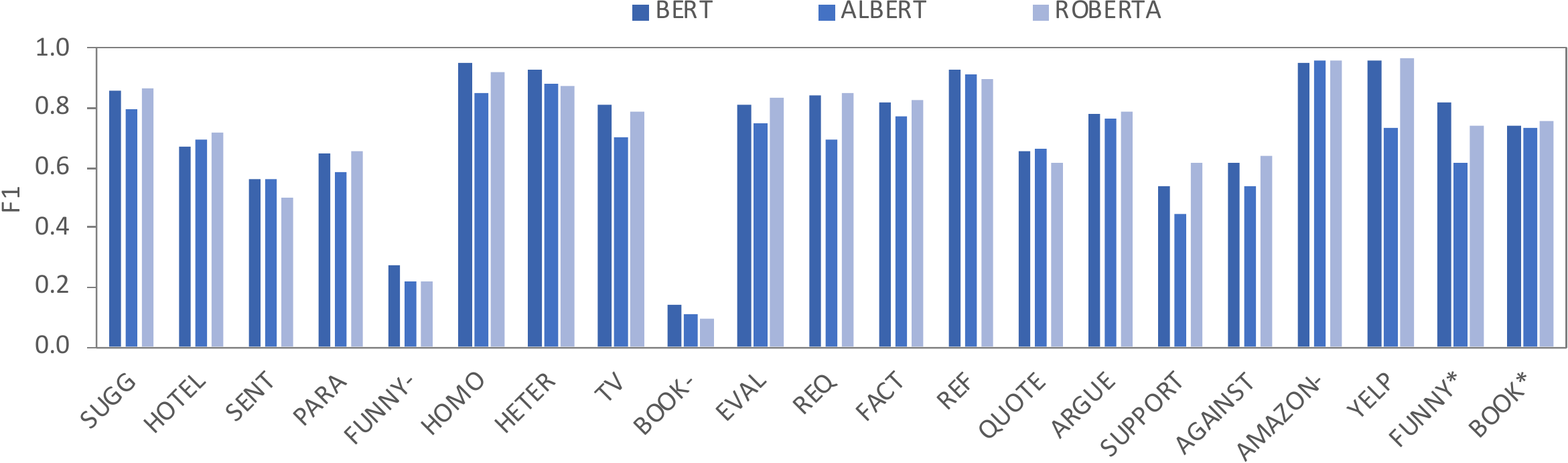}
        \caption{F1 of BERT, ALBERT, and ROBERTA}
	    \label{fig:albert_roberta}	
\end{figure*}

We conducted experiments to evaluate how much simple models can benefit from pre-trained embeddings. Given that BERT is similar to Word2Vec and Glove that use word vectors, for the convenience, we used BERT that was pre-trained over Wikipedia corpus as the embedder. For each input text, BERT outputs the last-layer [CLS] vector~\cite{devlin2018bert} as the featurization vector. We then ran LR/SVM on all featurization vector and presented the F1 of LR and SVM in Figure~\ref{fig:pre_train_lr} and Figure~\ref{fig:pre_train_svm}, respectively. Figure~\ref{fig:pre_train_lr} shows that LR can achieve better F1s on small datasets with pre-training embeddings. The F1 improvements are the most obvious on three datasets: HOMO, HETER, and QUOTE. Similarly, SVM also achieves the biggest F1 improvements on these three datasets ( Figure~\ref{fig:pre_train_svm}). These results suggest that simple models can benefit from pre-trained embeddings. 

% XGBoost and Naive Bayes
\section*{Performance of More Models} 
%In our previous experiments, we choose LR and SVM as representative simple models, and BERT as a representative deep model. However, there are a variety of simple models and deep models. 

In this section, we investigate the performance of more types of simple and deep models other than LR, SVM, and BERT. The newly investigated simple models include Naive Bayes~\cite{naive_bayesian} and XGBoost~\cite{ChenG16kdd} (a ensemble/boosting model). The newly investigated deep models include ALBERT~\cite{lan2019albert} and ROBERTA~\cite{roberta}, which are obtained from Huggingface transformers~\cite{huggingface}. We choose these models because they gain great industrial success. 

We evaluate the performance of Naive Bayes and XGBoost on all 21 datasets and report their F1s in Figure~\ref{fig:naive_bayes_xgb}. We also include the best F1 of LR and SVM (denoted as LR/SVM) in the same figure for comparison. Naive Bayes obtains better F1 than LR/SVM on 4 datasets (HOMO, EVAL, REF, and QUOTE), and similar or worse F1 on the remaining 17 datasets. XGBoost achieves better F1 than LR/SVM on 2 datasets (REQ and QUOTE), and similar or worse F1 on the remaining 19 datasets. The average F1 of Naive Bayes, XGBoost, and LR/SVM are 0.62, 0.61, and 0.65, respectively. Based on the results, LR/SVM achieves the best average F1. Therefore, we use LR and SVM as the representative simple models in our study. 

We evaluate the performance of ALBERT and ROBERTA on all 21 datasets. For FUNNY, BOOK, and AMAZON, we randomly sample 400k labels to ensure the model training can terminate within 24 hours. The sampled datasets are denoted as FUNNY-, BOOK-, and AMAZON-, respectively. We report F1s of ALBERT and ROBERTA in Figure~\ref{fig:albert_roberta}. We also include BERT in the figure for comparison. ALBERT outperforms BERT on HOTEL, and shows similar or worse F1 on the remaining 20 datasets. ROBERTA achieves better F1 than BERT on HOTEL and SUPPORT datasets, and shows worse F1 on SENT, HOMO, HETER, TV, BOOK-, REF, QUOTE and FUNNY*. The average F1 of ALBERT, ROBERTA, and BERT is 0.68, 0.72, and 0.73, respectively. Based on the results, BERT achieves the best average F1. Therefore, we take BERT as the representative attention-based deep model in our study. 

\section*{Performance on More Evaluation Measures}

We conduct extensive experiments to obtain Accuracy and AUC (area under the ROC curve)~\cite{area_under_roc_curve} scores, which are complements to F1. Specifically, we evaluate the performance of LR, SVM, CNN, LSTM, and BERT on all 21 datasets. For FUNNY, BOOK, and AMAZON datasets, we randomly sample 400,000 labels for each dataset to ensure that deep models can finish training within 24 hours. The sampled datasets are denoted as FUNNY-, BOOK-, and AMAZON-. To organize Accuracy and AUC results, we group 21 datasets into 4 categories by dataset characteristics (Table~\ref{table:category}), which are consistent with our reporting of F1 in Figure~\ref{fig:f1_high_p} and Figure~\ref{fig:f1_low_p}.

\begin{figure*}[]
    \centering
    \scalebox{1}{
    \includegraphics[width=\linewidth]{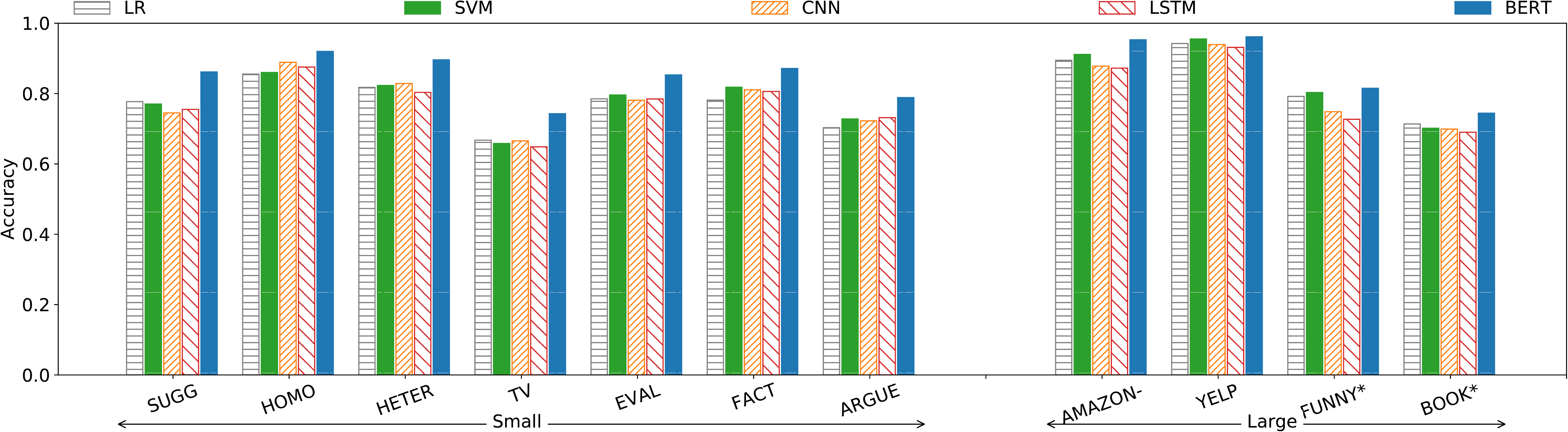}
    }
    \caption{Accuracy on \smallh\ and \largeh\ datasets with $\geqslant$ 25\% positive labels}
    \label{fig:accuracy_high_p}
\end{figure*}

\begin{figure*}[t]
    \centering
    \scalebox{1}{
    \includegraphics[width=\linewidth]{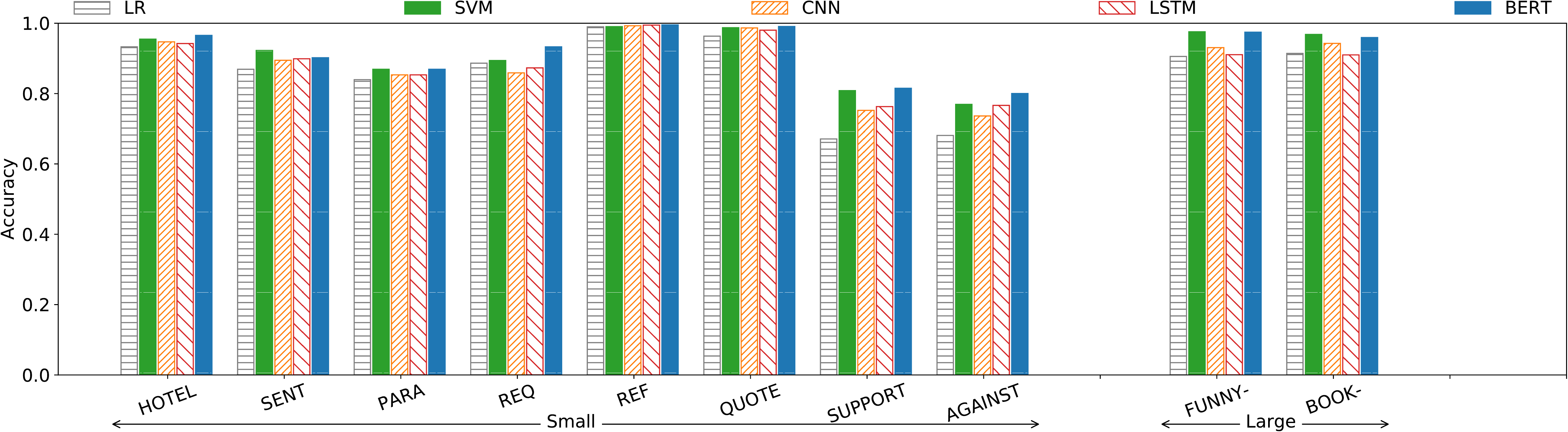}
    }
    \caption{Accuracy on \smalll\ and \largel\ datasets with $<$ 25\% positive labels}
    \label{fig:accuracy_low_p}
\end{figure*}

\begin{figure*}[t]
    \centering
    \scalebox{1}{
    \includegraphics[width=\linewidth]{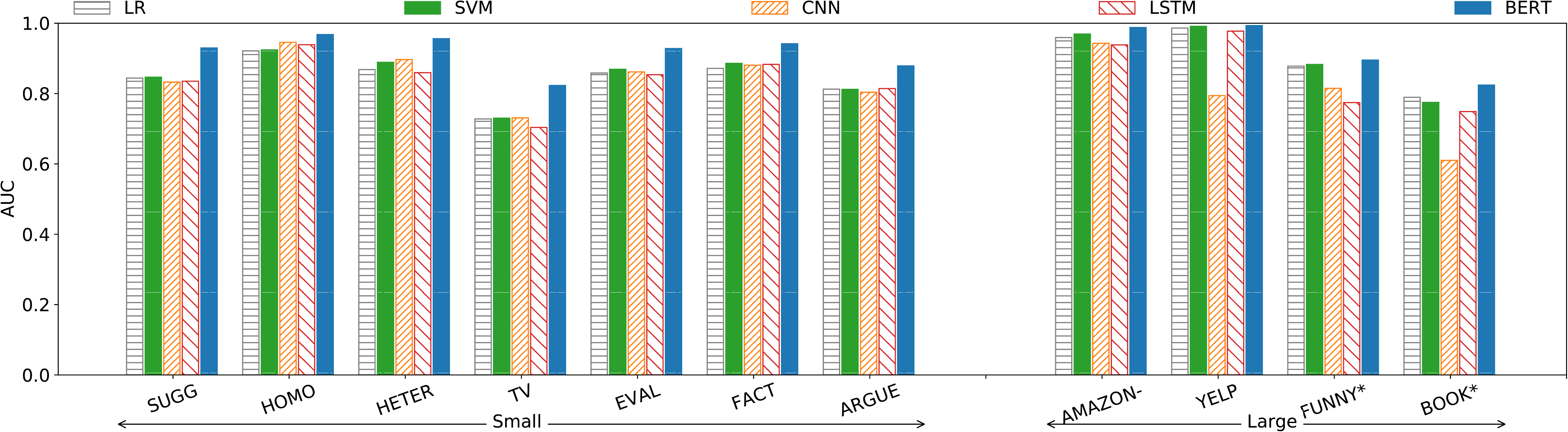}
    }
    \caption{AUC (Area under ROC curve) on \smallh\ and \largeh\ datasets with $\geqslant$ 25\% positive labels}
    \label{fig:auc_high_p}
\end{figure*}

\begin{figure*}[t]
    \centering
    \scalebox{1}{
    \includegraphics[width=\linewidth]{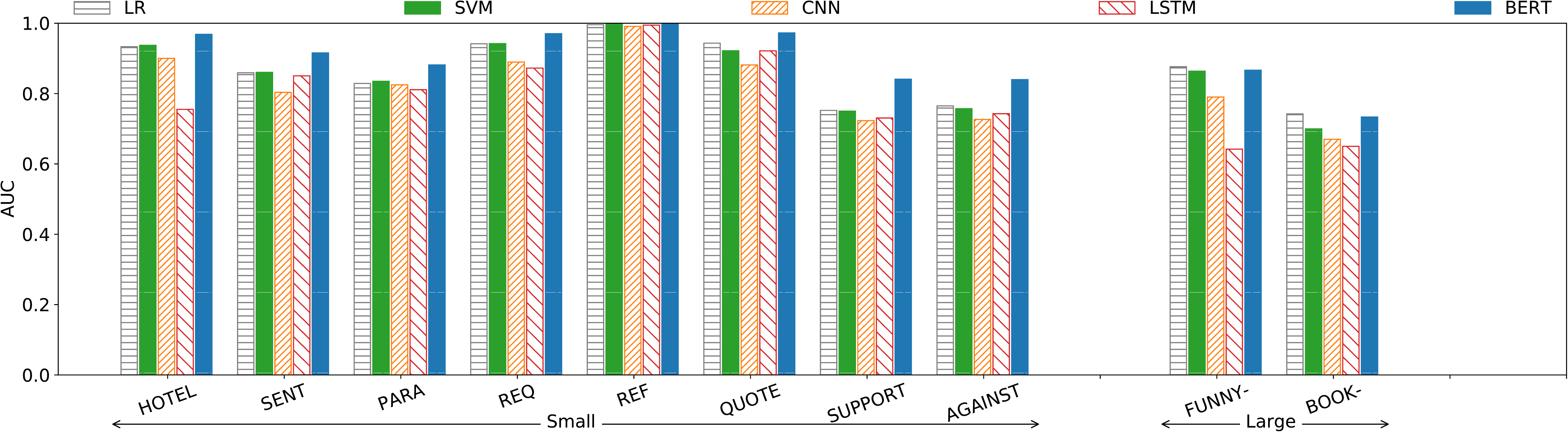}
    }
    \caption{AUC (Area under ROC curve) on \smalll\ and \largel\ datasets with $<$ 25\% positive labels}
    \label{fig:auc_low_p}
\end{figure*}

Accuracy results are presented in Figure~\ref{fig:accuracy_high_p} and Figure~\ref{fig:accuracy_low_p}, and AUC results are presented in Figure~\ref{fig:auc_high_p} and Figure~\ref{fig:auc_low_p}. Figure~\ref{fig:accuracy_high_p} and Figure~\ref{fig:auc_high_p} show datasets with $ \geqslant 25\%$ positive labels and Figure~\ref{fig:accuracy_low_p} and Figure~\ref{fig:auc_low_p} show datasets with $< 25\%$ positive labels. In comparison to F1 measurement, Accuracy and AUC scores do not show a clear tendency in the effect of label ratio: the higher, the better. For example, QUOTE has only 1.6\% positive labels but its Accuracy scores on LR, SVM, CNN, LSTM, and BERT are as high as 0.96, 0.99, 0.99, 0.98, and 0.99. Similarly, the AUC scores of LR, SVM, CNN, LSTM, and BERT are as high as 0.94, 0.92, 0.88, 0.88, and 0.94 on QUOTE. We suspect the low correlation of label ratio and performance revealed by Accuracy and AUC are due to that Accuracy and AUC are influenced by positives and negatives simultaneously. So Accuracy and AUC should be applicable for evaluations of multiple labels.  But such influences on F1 are different, since F1 is only influenced by positives. Therefore, F1 is more suitable for the evaluation of tagging quality of a targeted label dedicatedly. For example, if a practitioner aims at “not funny”, s/he can treat “not funny” as the targeted label. S/he can perfectly use our heat map through the label size, the label ratio, and the label cleanliness of “not funny”. Considering that in our study, we focus on evaluating a single label, perhaps F1 is a better choice for the default measure. 


\section*{Extension to NER and Knowledge Extraction}
We tentatively extend our experimental study to Named Entity Recognition (NER) and Knowledge Graph Extraction. We inspect SEMEVAL2020 competition~\cite{semeval2020} to look for related datasets. From task 6 of SEMEVAL2020, we found two datasets, BIO and DEF, that are relevant. BIO is a NER dataset while DEF is a Knowledge Graph Extraction dataset. %BIO contains around 470, 000 labels so the dataset is large. Each label is associated with a word token, and is either B, I or O. The best performing simple and deep models show similar F1s (same regarding labels B and O, 0.02 difference regarding label I). The results support our finding that simple models can achieve similar performance as deep models when the number of labels is sufficiently large. DEF contains around 18, 000 labels so the dataset is small. Each label is associated with a sentence, and is either T or F. The best performing deep model outperforms the best simple model on F1 (0.14 gap regarding label T and 0.04 improvement regarding label F). The results support our finding that deep models achieve better performance than simple models when the number of labels is small. 

BIO contains around 470, 000 labels so the dataset is large. Each label is associated with a word token and is either B, I or O. B/I/O represents a token is the begin/inside/outside of a targeted token sequence. These labels are used to train models to identify token sequences that express term definitions from free texts~\cite{task6_semeval2020}. Since there are three classes (B, I, and O), we evaluate LR, SVM, CNN, LSTM, and BERT as a three-class classification. The F1s regarding label B are 0.01, 0.08, 0.04, 0.08, and 0.08, respectively. The F1s regarding label I are 0.07, 0.13, 0.06, 0.15, and 0.13, respectively. The F1s regarding label O are 0.85, 0.85, 0.85, 0.85, 0.85, respectively. The best performing simple/deep model is SVM/LSTM. The F1 scores of SVM and LSTM are very similar (same regarding labels B and O, 0.02 difference regarding label I). The results support our finding that simple models can achieve similar performance as deep models when the number of labels is sufficiently large. %\xiaolan{For BIO tagging task, people typically do not evaluate O labels. For B/I tags, the numbers look extremely low (even for BERT), which is a bit strange.}

DEF contains around 18, 000 labels so the dataset is small. Each label is associated with a sentence and is either T or F that indicates whether the sentence contains a term definition~\cite{task6_semeval2020}. These labels are used to train models to identify sentences that express term definitions from free texts. We evaluate LR, SVM, CNN, LSTM, and BERT as a binary classification. The F1 regarding label T is 0.72, 0.72, 0.68, 0.66, and 0.80. The F1 regarding label F is 0.86, 0.83, 0.84, 0.85, and 0.90. The best performing simple/deep model is LR/BERT. BERT outperforms LR on F1 (0.14 gap regarding label T and 0.04 gap regarding label F). The results support our finding that deep models achieve better performance than simple models when the number of labels is small. 

%NER results show preliminary comparison of simple and deep models on the setting of token-level tagging. This is different from text-level tagging which is the main focus of current manuscript. Knowledge Graph Extraction offers more results in addition to current results over 21 datasets. Specially, Knowledge Graph Extraction presents F1s of both T and F classes, while current results report F1s of the targeted label only (by default T) to understand the effect of label ratio. The NER and Knowledge Graph Extraction results serve as great reference for future research to compare simple and deep models in new settings that concern token-level tagging or multiple-label evaluation. 

NER results show a preliminary comparison of simple and deep models on the setting of token-level tagging. This is different from text-level tagging which is the main focus of current manuscript. Knowledge Graph Extraction offers more results in addition to current results over 21 datasets. Specifically, Knowledge Graph Extraction presents F1s of both T and F classes, while current results report F1s of the targeted label only (by default T) to understand the effect of label ratio. The NER and Knowledge Graph Extraction results serve as a great reference for future research to compare simple and deep models in new settings that concern token-level tagging or multiple-label evaluation. 

%\section*{Comparison on Multi-class Classification} 
% Experiments on multi-class classification confirms out finding. For a targeted class, deep models does not outperform simple models since the extraction F1 is affected by size, label ratio, and cleanliness. 

\begin{table}[t]
\centering
\begin{tabular}{c|ccccc}
\toprule
 & LR & SVM & CNN & LSTM & BERT \\ \midrule
\largeh & 0.77 & 0.76 & 0.72 & 0.72 & 0.79 \\ \midrule 
\smallh & 0.73 & 0.72 & 0.70 & 0.71 & 0.82 \\ \midrule
\smalll & 0.51 & 0.51 & 0.47 & 0.49 & 0.66 \\ \midrule
\largel & 0.20 & 0.20 & 0.06 & 0.11 & 0.19 \\ \bottomrule
\end{tabular}
\caption{The micro average F1 of LR, SVM, CNN, LSTM, and BERT in dataset categories}
\label{table:avg_micro_f1_category}
\end{table}

\section*{Overall Comparison with Micro F1}
To compare overall performance of simple and deep models, we report macro F1 in Table~\ref{table:avg_f1_category} and Figure~\ref{fig:avg_f1}. Macro F1 is the average F1 scores, for which it is insensitive to the sizes of datasets. Given that the 21 datasets are of various sizes, we calculate micro-average F1, the sum of weighted F1s, as a complement to macro-average F1. Specifically, the weight of a dataset is the number of records of this dataset divided by total number of records of all datasets. Therefore, a larger dataset will have a higher weight than a smaller dataset.  

%We report the micro-average F1 scores of LR, SVM, CNN, LSTM, BERT regarding \largeh, \smallh, \smalll, and \largel\ datasets in Table 8. LR and SVM show worse scores than BERT on small datasets. The gap is 0.09/0.15 on \smallh/\smalll, respectively. However, LR and SVM achieve similar scores as BERT on large datasets. LR shows 0.02 lower F1 on \largeh\ and 0.01 higher F1 on \largel. SVM shows 0.03 lower F1 on \largeh\ and 0.01 higher F1 on \largel. The results suggest that simple models can achieve similar tagging performance as deep models when datasets are large. This is consistent with our macro-average F1 results in Table 5. 

%Next we report the micro-average F1 scores over all 21 datasets. The score of LR/SVM/CNN/LSTM/BERT is 0.33/0.34/0.22/0.25/0.33, respectively. LR achieves the same score as BERT while SVM slightly outperforms BERT by 0.01. The results indicate simple models achieve overall the same performance as deep models on all 21 datasets. This is because micro-average F1 is significantly dominated by large datasets. The weights of the 6 large datasets sum up to 0.99, while weights of the 15 small datasets sum to 0.01. Besides, the micro-average F1 is significantly dominated by BOOK, the largest dataset. The weight of BOOK is 0.63. Since micro-average F1 is biased towards large datasets, we do not use it as the default measure in our paper. 

We reported the micro-average F1 scores of LR, SVM, CNN, LSTM, and BERT regarding Large-H, Small-H, Small-L, and Large-L datasets in Table 8. The results suggest that simple models achieve similar tagging performance as deep models on large datasets (i.e. Large-H and Large-L). This is consistent with our macro-average F1 results in Table 5. Next, we reported the micro-average F1 scores over all 21 datasets. The score of LR, SVM, CNN, LSTM, and BERT is 0.33, 0.34, 0.22, 0.25, and 0.33, respectively. The results indicate that simple models achieve overall the same performance as deep models on all 21 datasets. This is because that micro-average F1 is dominated by the performance of large datasets, where simple and deep models achieve similar performance. The weights of the 6 large datasets sum up to 0.99, while weights of the remaining 15 small datasets sum to 0.01. Therefore, simple models are indeed as competitive as deep models.